\documentclass[letterpaper, 10pt, conference]{ieeeconf}

\IEEEoverridecommandlockouts 
\usepackage{fancyhdr}
\usepackage{graphicx}
\usepackage{amsmath}
\usepackage{amssymb}
\usepackage{booktabs}
\usepackage{array}
\usepackage{multirow}
\usepackage{subcaption}
\usepackage{hyperref}
\hypersetup{hidelinks}
\usepackage{xcolor}
\usepackage{url}
\usepackage{listings}
\usepackage{float}
\usepackage{placeins}
\usepackage{pifont}

\lstdefinestyle{yaml}{ basicstyle=\ttfamily\fontsize{6.2}{7.2}\selectfont, columns=fixed,
  backgroundcolor=\color{gray!10}, frame=none, breaklines=true,
  showstringspaces=false, xleftmargin=1mm, xrightmargin=1mm,
  keepspaces=true, captionpos=b }

\newcommand{\legendswatch}[1]{%
  \raisebox{0.25ex}{%
    {\setlength{\fboxsep}{0pt}\colorbox[HTML]{#1}{\rule{0pt}{1.0ex}\rule{2.0ex}{0pt}}}%
  }%
}

\title{\LARGE \bf SLAM Adversarial Lab: An Extensible Framework for Visual SLAM Robustness Evaluation under Adverse Conditions
}

\author{Mohamed Hefny$^{1}$, Karthik Dantu$^{2}$, Steven Y. Ko$^{1}$%
\thanks{$^{1}$Simon Fraser University. \{mohamed\_hefny, steveyko\}@sfu.ca}%
\thanks{$^{2}$University at Buffalo. kdantu@buffalo.edu}%
}

\begin{document}

\maketitle
\thispagestyle{fancy}
\fancyhf{}
\fancyfoot[C]{\footnotesize This work has been submitted to the IEEE for possible publication. Copyright may be transferred without notice, after which this version may no longer be accessible.}
\bstctlcite{BSTcontrol}

\begin{abstract}
We present SAL (SLAM Adversarial Lab), a modular framework for evaluating visual SLAM
systems under adversarial conditions such as fog and rain.
SAL represents each adversarial condition as a \emph{perturbation} that transforms
an existing dataset into an adversarial dataset. When
transforming a dataset, SAL supports severity levels using easily-interpretable
real-world units such as meters for fog visibility. SAL's extensible architecture
decouples datasets, perturbations, and SLAM algorithms through common interfaces,
so users can add new components without rewriting integration code. Moreover,
SAL includes a search procedure that finds the severity level of a perturbation
at which a SLAM system fails. To showcase the capabilities of SAL, our evaluation
integrates seven SLAM algorithms and evaluates them across three datasets under
weather, camera, and video transport perturbations.
\end{abstract}

\section{Introduction}
\label{sec:introduction}

\noindent Visual SLAM systems achieve high accuracy on clean benchmark
sequences~\cite{bujanca2021robustslam}, yet real-world deployments expose them to
conditions that fall outside the assumptions behind this accuracy.
Adverse weather and illumination (rain, fog, low light), camera degradations
(lens soiling, cracking, motion blur), and video transport effects (compression,
frame drops) can break the photometric and geometric assumptions that visual SLAM
relies on, reducing trackability and sometimes causing complete loss of
localization~\cite{Wenzel2025}.
In safety-critical applications such as autonomous driving, such failures
motivate adversarial robustness evaluation that is controlled, reproducible,
and comparable across algorithms.

We identify three practical needs for meaningful adversarial robustness evaluation in visual SLAM. 
First, \textit{operational interpretability}: adverse conditions should be parameterized in real-world units 
(e.g., fog visibility in meters, rain intensity in mm/h), rather than abstract labels such as ``corruption severity 3/5'' or ``noise scale 50\%.''
Second,
\textit{modular extensibility}: degradation effects, datasets, and SLAM algorithms should be
modular and pluggable so that adding or removing a component does not require modifying the others.
Third,
\textit{failure-boundary localization}: evaluation should be able to automatically determine at which severity level SLAM performance transitions from pass to fail. In addition, a user
should be able to define a failure threshold (e.g. ATE exceeding 1.5\,m), instead of testing only a few fixed severity levels.

However, current approaches for adversarial robustness evaluation in visual SLAM do not satisfy these needs. Typically, researchers either (i)
collect new datasets under desired conditions~\cite{wenzel20204seasons}, or
(ii) write \emph{perturbation} code to add an adverse condition specifically to a dataset or a SLAM system~\cite{cao2020ffdnet,sarti2023robust}. Collecting new datasets captures only the conditions present during recording, so it offers no way to (i)
vary degradation severity (e.g., 3\,mm/h vs.\ 200\,mm/h rain) or (ii) isolate a single degradation source or compose multiple degradation
sources beyond whatever happened to co-occur at fixed severity in the recorded
dataset. For example, a recorded fog sequence may not match a target visibility such as
10\,m and may include unintended coupled effects such as night-time darkening,
whereas starting from clean data allows applying fog alone or combining fog
with other degradations in a controlled way. The second approach,
writing perturbation code for specific datasets or SLAM algorithms, tightly couples implementation to dataset conventions
(e.g., directory layouts, image naming schemes) and algorithm-specific requirements (e.g., trajectory output format). As a result, reusing the same perturbation on a different dataset or SLAM algorithm,
or introducing a new perturbation, often requires nontrivial code rewrites. Additionally,
neither approach provides a systematic way to identify the exact perturbation severity at which a SLAM system fails.

To address the three practical needs and the limitations of existing approaches, we propose SAL (SLAM Adversarial Lab),
a modular framework for evaluating visual SLAM robustness under adversarial
conditions. To ensure realistic condition modeling, SAL uses perturbation models
with interpretable parameters (e.g., fog visibility in meters) that are
depth-aware and support combining multiple perturbations (e.g., rain + motion blur), rather than simple image corruptions.
To support extensibility, SAL separates datasets, perturbations, and SLAM
algorithms so that each can be added or replaced independently.
To enable fine-grained evaluation, SAL includes a search procedure that
estimates the perturbation severity at which a SLAM system fails.

We demonstrate SAL capabilities by evaluating seven SLAM algorithms
across three datasets covering outdoor and indoor
scenes with monocular, stereo, and RGB-D configurations. Our adversarial evaluations show that composite perturbations
such as night combined with fog cause significantly larger degradation than
individual effects, and that algorithms vary widely in their sensitivity
to different perturbation types. Feature tracking diagnostics show that degradation can be abrupt:
ORB~\cite{murORB2} features track stably through heavy rain (50 mm/h) but fail completely at severe
intensity (200 mm/h).
The search procedure pinpoints failure thresholds that
testing at a few fixed severity levels misses, for example, for ORB-SLAM3~\cite{Campos2021ORBSLAM3} on
KITTI~\cite{Geiger2012CVPR} night+fog, a 3\,m visibility change causes about a 93-fold increase in
ATE RMSE, showing that performance can degrade sharply within a narrow
visibility range.

\section{Related Work}
\label{sec:related_work}
\noindent This section reviews related work on SLAM evaluation and perturbation frameworks,
adverse-condition datasets, and robust SLAM algorithms.

\subsection{SLAM Evaluation Frameworks}
\label{subsec:evaluation_frameworks}
\noindent SLAM evaluation frameworks simplify experiment execution and make results more reproducible and comparable.
SLAMBench~\cite{nardi2015slambench} provided a common way to run SLAM algorithms on datasets and report accuracy alongside system cost such as runtime and energy.
VSLAM-LAB~\cite{fontan2025vslamlab} offers a unified command-line workflow that streamlines algorithm compilation and configuration, automates dataset preparation, and standardizes experiment execution and evaluation.
These frameworks improve standardized SLAM execution and modular integration, but
they do not target controlled and extensible adversarial robustness evaluation.

Other work evaluates robustness by generating perturbed inputs and measuring degradation and failure behavior.
SLAMFuse~\cite{radulov2024slamfuse} adds systematic dataset perturbations and diagnostics to identify where performance breaks down,
but focuses on basic image manipulations such as brightness, contrast, and blur, rather than modeling structured, deployment-realistic effects. 
Moreover, it does not provide a modular plug-in interface for perturbations, requiring core framework changes to introduce new effects.
Robust-Ego3D~\cite{xu2024robustslam} synthesizes perturbed benchmarks by rendering clean 3D scenes and combining motion, imaging, depth, and synchronization perturbations, with an indoor RGB-D benchmark based on Replica scenes.
In contrast, SAL applies parameterized perturbations in interpretable, real-world units to
indoor and outdoor sequences across monocular, stereo, and RGB-D
setups. Its architecture is extensible across datasets, perturbation modules,
and SLAM algorithms, supports composite visual perturbations, and includes a search
procedure for estimating failure thresholds.

\subsection{Adverse Condition Datasets}
\label{subsec:adverse_conditions}
\noindent Several datasets capture adverse conditions and highlight robustness gaps beyond
clean benchmarks. Outdoor examples include Oxford
RobotCar~\cite{RobotCarDatasetIJRR} with rain, night, direct sunlight, and snow,
4Seasons~\cite{wenzel20204seasons} with seasonal, weather, and illumination
variation, and SubT-MRS~\cite{Zhao2024} with underground and degraded visibility
scenes. Indoors, QueensCAMP~\cite{Bruno2024} includes motion blur, lighting
changes, and simulated camera failures such as dirt, condensation, and
under/over-exposure. TartanAir~\cite{Wang2020} provides photo-realistic
simulation with weather and lighting variation.

While these datasets provide pre-collected adverse conditions, SAL
applies controlled, parameterized perturbations to existing sequences.
This enables clean-vs-perturbed comparisons on identical scenes, reproducible
evaluation across severity levels, and composite perturbation experiments without
costly and time-consuming data collection for each operating condition.

\subsection{Robust SLAM Algorithms}
\label{subsec:robust_algorithms}
\noindent Robust SLAM methods aim to reduce sensitivity to adverse visual conditions such as illumination change, weather, and reduced visibility. NID-SLAM~\cite{Pascoe2017NIDSLAM} replaces photometric error with an information-theoretic objective, improving robustness to illumination variation and related appearance changes. Other work targets low-light operation by applying image enhancement before a conventional SLAM pipeline, as in TwilightSLAM~\cite{Singh2023TwilightSLAM} and related low-light ORB-SLAM3 variants~\cite{Han2024LowLightORBSLAM3}. Unlike these methods, our work does not propose a new SLAM algorithm. Instead, we provide a framework to systematically evaluate SLAM algorithms under controlled, parameterized adverse conditions.

\section{SAL (SLAM Adversarial Lab)}
\label{sec:architecture}

\noindent The framework is designed to meet three
evaluation needs. First, operationally meaningful parameterization in units
that map to real-world operating conditions.
Second, modular architecture in
which datasets, perturbation modules, and SLAM algorithms are decoupled and pluggable,
so each can be reused across configurations without changing the others. Third, systematic identification of
failure thresholds beyond a few fixed severity levels. To meet these needs, we make three
contributions through SAL that support systematic studies of robustness under
deployment-relevant conditions:

\textbf{(C1) Specifying realistic conditions:}
we define adverse conditions as perturbation modules which are parameterized in real-world
units, such as fog visibility in meters or rain intensity in mm/h.
\textbf{(C2) Modular and adaptable:}
SAL is built around three pluggable extension points: dataset adapters, perturbation modules, and SLAM wrappers. 
Each component follows a standard interface, so new datasets, algorithms, 
or perturbations can be added without modifying other framework components.
\textbf{(C3) Test to failure:}
we introduce a bisection-based search that estimates the perturbation
severity at which a SLAM system fails, where failure is defined by a
user-specified metric threshold (e.g., ATE RMSE exceeding 1.5\,m). The remainder of this section details the contributions (Sections~\ref{subsec:modules}--\ref{subsec:robustness_boundary}) and experiment execution pipelines (Section~\ref{subsec:experiment_execution}).

\subsection{Perturbation Modules}
\label{subsec:modules}


\noindent To support (C1), SAL defines each adverse condition as a
perturbation module. We aim to model degradations as realistically as possible while keeping robustness conclusions relevant to deployment. Accordingly, we define
three requirements for realistic
adversarial robustness evaluation: (i) perturbations should model specific operating
conditions with condition-relevant parameters (e.g., fog visibility in m, rain
intensity in mm/h) so robustness outcomes map to real operating conditions,
rather than generic pixel-level corruptions not tied to a particular condition, (ii)
they should support depth-aware parameterization when the effect depends on scene
geometry so the perturbation effect varies with distance, as in real scenes, and (iii)
they should be composable because real conditions can involve multiple simultaneous
degradations whose interaction can be more harmful than any single effect.
SAL addresses each of these requirements as follows:

\textbf{Perturbation realism:}
A key goal is to model deployment-relevant degradations rather than generic
photometric corruptions. Where applicable, modules expose parameters in
interpretable, real-world units rather than abstract corruption levels. For example,
we parameterize \texttt{rain} by intensity in mm/h, \texttt{fog} by visibility
in meters, and \texttt{motion\_blur} by camera speed (km/h) and exposure
time (ms). We relate rain and fog values to standard meteorological
categories~\cite{ams_rain,metoffice2007,metoffice_fog}.

\textbf{Depth-aware parameterization:}
Real-world degradations are depth-dependent, so perturbation strength is conditioned on scene depth and near/far regions are affected differently (e.g., fog transmission, rain attenuation, motion blur). For instance, with fog visibility set to 50\,m, distant content is affected more than nearby content, and content at 50\,m and beyond tends to blend into the fog. The framework first reuses existing depth maps, including native RGB-D depth. If depth is missing and stereo calibration is available, it computes metric depth from stereo pairs using FoundationStereo~\cite{wen2025stereo} by converting disparity with calibration. Otherwise, it estimates monocular depth with Depth Anything V2~\cite{depthanything}. Depth is cached per sequence and reused across modules to avoid redundant computation.

\textbf{Composite perturbations:}
Real deployments often involve multiple simultaneous conditions.
SAL supports composite perturbations that combine multiple effects in order,
for example, rain with lens soiling or fog with motion blur.
Users can compare each effect in isolation against its combination on the same
sequence to determine whether the combined degradation exceeds either individual
effect.

\subsection{Extensible Benchmarking Architecture}
\label{subsec:extensible_architecture}

\noindent To support extensible studies (C2), users should be able to integrate new datasets,
perturbations, and SLAM algorithms as modular components through well-defined extension points,
then reuse them across many benchmark configurations (dataset $\times$ perturbation $\times$ SLAM algorithm).
Each new component should integrate once at its extension point, without requiring changes elsewhere
in the framework. This matters because robustness studies require many combinations,
and tightly coupled integrations increase implementation effort, slow experimentation,
and lead to inconsistent comparisons. SAL provides three extension points for this: dataset adapters,
perturbation modules, and SLAM wrappers. The following paragraphs describe each extension point with examples.

\textbf{Datasets:} Each dataset adapter implements a common interface
that specifies, among other methods, where to find the dataset sequence images, how
frames are ordered in time so that perturbations remain temporally coherent,
and where depth maps are located if the dataset provides them. For example, the
KITTI~\cite{Geiger2012CVPR} adapter points to files in \texttt{image\_2/} and \texttt{image\_3/} for
stereo and orders frames by filename index. The TUM~\cite{sturm12iros} adapter instead points to \texttt{rgb/},
orders frames by timestamp, and exposes native RGB-D depth maps directly. Because each adapter maps these details to the common interface,
perturbation modules and SLAM wrappers work identically on any dataset
without modification. Supporting a new dataset requires implementing an adapter
that maps its file layout to this interface.

\textbf{Perturbations:} Each perturbation module implements three methods:
\texttt{setup()}, \texttt{apply()}, and \texttt{cleanup()}. The \texttt{setup()} method receives the dataset context which
includes the dataset path and total frame count, and prepares any resources
the module needs. The \texttt{apply()} method receives each frame's image,
depth map, frame index, and camera stream (e.g., left or right in stereo), and
returns the perturbed image.
\texttt{cleanup()} frees any resources the module allocated. For example, the
fog module's \texttt{setup()} generates depth maps for the full sequence when
the dataset does not provide native depth. Its \texttt{apply()} then loads each
pre-computed depth map and applies the Koschmieder scattering model per pixel,
attenuating distant features more than nearby ones. \texttt{cleanup()} releases
the depth model.

\textbf{SLAM algorithms:} Each SLAM wrapper implements a common interface
that stages the dataset
into the layout expected by the algorithm,
resolves dataset/sequence-specific settings when required by that algorithm, executes it, and
converts the resulting trajectory to a common TUM format so that the evaluation
pipeline can compute metrics regardless of which algorithm produced it.
For example, the ORB-SLAM3 wrapper resolves
which ORB-SLAM3 camera-settings file to use by mapping the dataset
name and sequence (e.g., KITTI sequence \texttt{"04"} to \texttt{KITTI04-12.yaml}).
It stages the original and perturbed stereo sequences by symlinking image directories
into ORB-SLAM3's expected stereo layout (e.g., \texttt{image\_2} to
\texttt{image\_0}), and executes the
algorithm. Because ORB-SLAM3 already outputs TUM format, no conversion is
needed. Wrappers for algorithms that use other formats, such as S3PO-GS~\cite{cheng2025s3pogs} which
outputs poses in JSON, convert the result to TUM format. To improve
reproducibility and reduce environment-specific setup issues, each algorithm
runs in its own runtime environment.

\subsection{Robustness Boundary Evaluation}
\label{subsec:robustness_boundary}

\noindent To support (C3), SAL estimates the parameter value at which SLAM first fails a criterion, e.g., finding the fog visibility at which ATE RMSE exceeds 1.5\,m.
This allows the estimated boundary to indicate when fallback actions may be needed, such as reducing speed or executing a controlled stop.
The rest of this section describes the search procedure and how perturbation modules declare searchable parameters, allowing
boundary evaluation to extend to new modules.


\subsubsection{Search Procedure}
\label{subsubsec:search_procedure}
\noindent Boundary search evaluates the target perturbation at selected
parameter values and marks each evaluation as pass or fail against a configured
ATE RMSE threshold.
It then narrows the parameter interval with bisection until the
interval width is within tolerance or the maximum number of iterations is
reached. As an illustrative example, in fog visibility search with a range of 10--20\,m,
failure defined as ATE RMSE $>$ 1.5\,m, and tolerance set to 2\,m, 
SAL first evaluates 10 m and 20 m. If the ATE is 2.3 m at 10 m, it is marked as fail because it exceeds 1.5 m, while 20 m (ATE = 0.8 m) is marked as pass because it does not exceed 1.5 m. SAL then tests 15 m, the midpoint between the current fail/pass values (10 m and 20 m), and marks it as pass (ATE = 1.1 m). It next tests 12 m (ATE = 1.4 m, pass). At that point, the current fail/pass pair is 10 m and 12 m, and 12 - 10 = 2 m, so the search stops and reports the failure boundary interval as [10 m, 12 m].
This reduces the number of evaluated
levels from linear in range size to logarithmic. 

\subsubsection{Extensibility}
\noindent Boundary evaluation builds on the perturbation module adapter interface. Modules declare
parameters the boundary search is allowed to vary
in a \texttt{SEARCHABLE\_PARAMS} map. Each declared
parameter can define an optional \texttt{canonicalize} hook so users can
specify boundary values in numeric form for search traversal, while the module
still receives values in its expected input format, e.g., bitrate \texttt{5} as
\texttt{"5M"} for
\texttt{NetworkDegradationModule} (Listing~\ref{lst:boundary_canonicalize}).
Users can reference these parameters in
the experiment YAML configuration used for experiment execution (Section~\ref{subsec:experiment_execution}), and the
robustness-boundary pipeline can search them without any additional module-specific search
logic.

\begin{lstlisting}[language=Python, basicstyle=\ttfamily\fontsize{6.2}{7.2}\selectfont, backgroundcolor=\color{gray!10}, captionpos=b, caption={Boundary parameter for the network module.}, label=lst:boundary_canonicalize]
class NetworkDegradationModule(PerturbationModule):
    module_name = "network_degradation"
    SEARCHABLE_PARAMS = {
        "target_bitrate": BoundaryParamSpec(
            domain="integer",
            canonicalize=lambda v: f"{int(v)}M"
        )
    }
\end{lstlisting}

\subsection{Experiment Execution}
\label{subsec:experiment_execution}

\begin{figure}[t]
\centering
\hspace*{-6mm}
\includegraphics[width=1.10\columnwidth,trim=6 6 6 0,clip]{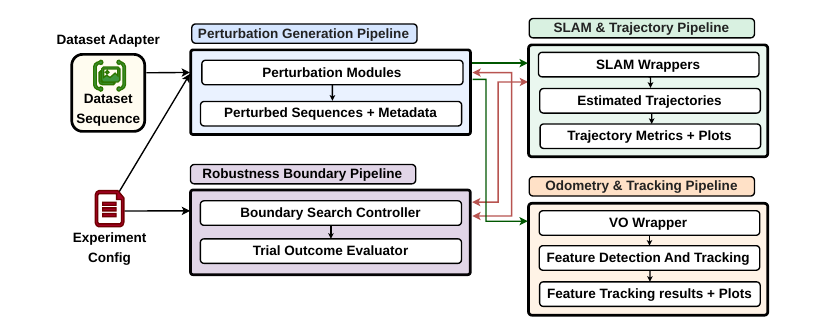}
\caption{System overview and data flow. Experiment configuration and dataset adapters feed the perturbation pipeline, which outputs perturbed sequences. The original and perturbed sequences are evaluated by the SLAM pipeline (trajectories, metrics, plots) and the odometry pipeline (tracking statistics). The robustness-boundary pipeline reuses perturbation and SLAM pipelines to localize the failure boundary.}

\label{fig:sys_overview}
\end{figure}

\noindent Users specify experiments through a declarative YAML configuration
that selects a dataset, sequence, and
perturbations. SAL provides four pipelines that users invoke separately
to generate perturbed data, execute SLAM and evaluate trajectories, diagnose feature
tracking, and search for robustness boundaries.
Figure~\ref{fig:sys_overview} summarizes this execution flow and the interfaces
between dataset adapters, perturbation modules, and SLAM wrappers.
The following paragraphs describe each pipeline.

\textbf{Perturbation generation.}
From a YAML experiment definition (Listing~\ref{lst:config}), this pipeline
loads the original
image sequence through the dataset adapter and applies each
configured perturbation module. 
The trajectory evaluation and feature
tracking pipelines can then evaluate different SLAM systems and feature
extractors on identical perturbed inputs without regenerating data.

\begin{lstlisting}[style=yaml, caption={Experiment configuration.}, label=lst:config]
experiment:
  name: kitti_fog_rain_soiling
dataset:
  type: kitti
  sequence: "07"
  max_frames: 200
  load_stereo: true       # perturb left and right images
perturbations:
  - name: fog_example     # outputs foggy sequence
    type: fog
    parameters:
      visibility_m: 100.0
  - name: soiling_rain    # outputs one sequence, both effects
    type: composite
    parameters:
      modules:
        - type: lens_soiling
          parameters:
            num_particles: 80
        - type: rain
          parameters:
            intensity: 100
output:
  base_dir: ./results/experiments
\end{lstlisting}

\textbf{SLAM execution and trajectory evaluation.}
For each selected SLAM algorithm, this pipeline runs its wrapper on the
clean and perturbed sequences, then computes Absolute Trajectory Error (ATE)
and Relative Pose Error (RPE) using evo~\cite{evo}. The pipeline can optionally run each algorithm N times and aggregate metrics across runs.
The output structure includes trajectories, metrics,
and plots (Listing~\ref{lst:slam_output}).

\begin{lstlisting}[caption={Trajectory evaluation output (RPE omitted).},label=lst:slam_output,basicstyle=\ttfamily\fontsize{6.2}{7.2}\selectfont,backgroundcolor=\color{gray!10},captionpos=b]
slam_results/<algorithm>/
  trajectories/run_<N>/
    baseline.txt          # baseline trajectory
    <perturbation>.txt    # perturbed trajectory
  metrics/
    run_<N>/
      baseline/
        ate.json          # absolute trajectory error
      <perturbation>/
        ate.json
        vs_baseline.json  # baseline vs perturbed summary
    comparison/run_<N>/
      ate_comparison.png  # ATE across perturbation settings
    summary.json          # aggregated statistics
    aggregated_metrics.png
  trajectory_plots/
    comparison/run_<N>/
      trajectory_comparison_*.png
\end{lstlisting}

\textbf{Odometry and feature tracking.}
Using PySLAM~\cite{freda2025pyslam}, this pipeline runs feature tracking on
the clean and perturbed sequences with a chosen extractor (e.g., ORB~\cite{murORB2}, SuperPoint~\cite{detone2018superpoint}), reporting statistics for each perturbation setting including mean track
length (the average number of frames a feature is successfully
matched), summary statistics of per-frame feature detections and matches, along
with track survival curves (\% of tracks surviving $\geq t$ frames).

\begin{figure*}[!t]
\vspace*{1.5mm}
\captionsetup[subfigure]{font=footnotesize,skip=1pt}
\centering
\begin{subfigure}[b]{0.195\textwidth}
    \includegraphics[width=\textwidth]{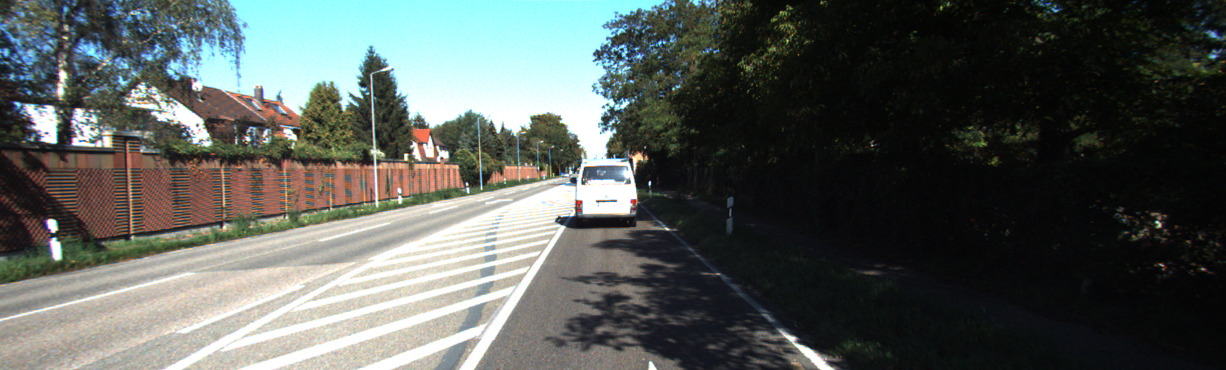}
    \caption{Original}
\end{subfigure}
\hfill
\begin{subfigure}[b]{0.195\textwidth}
    \includegraphics[width=\textwidth]{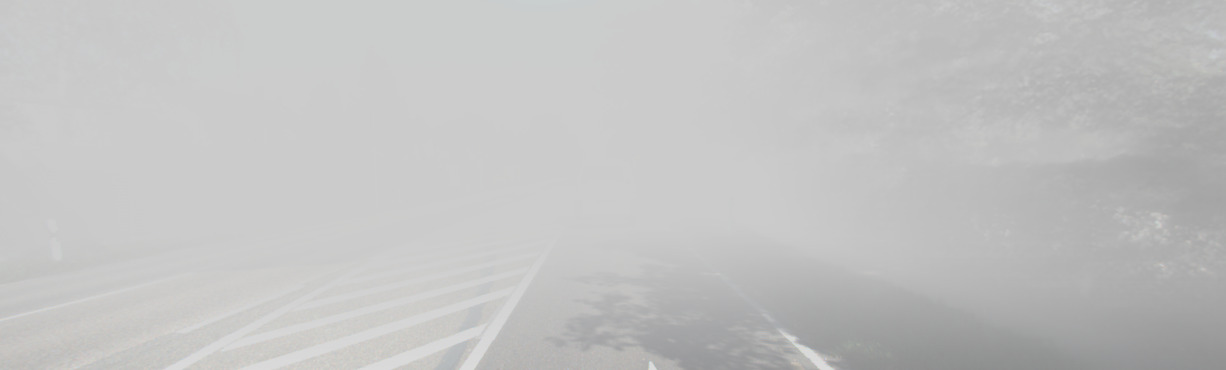}
    \caption{\mbox{Fog (Heavy)}}
\end{subfigure}
\hfill
\begin{subfigure}[b]{0.195\textwidth}
    \includegraphics[width=\textwidth]{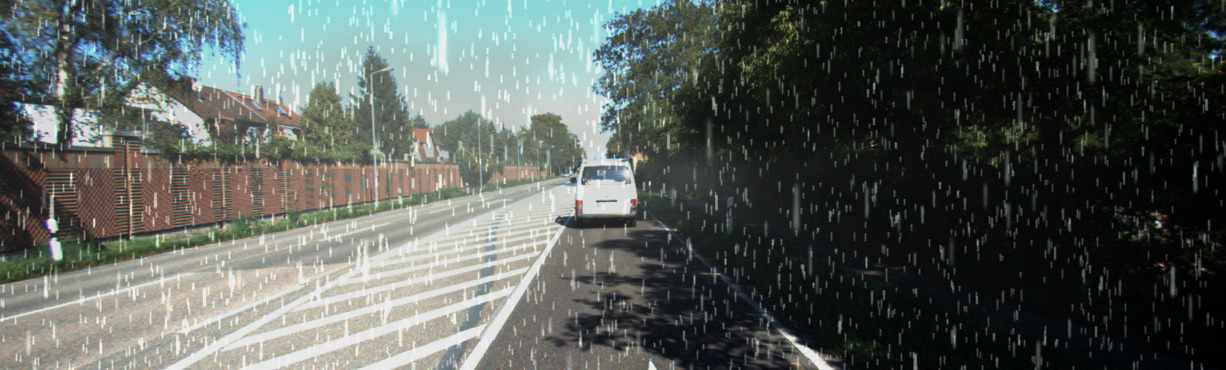}
    \caption{\mbox{Rain (Severe)}}
\end{subfigure}
\hfill
\begin{subfigure}[b]{0.195\textwidth}
    \includegraphics[width=\textwidth]{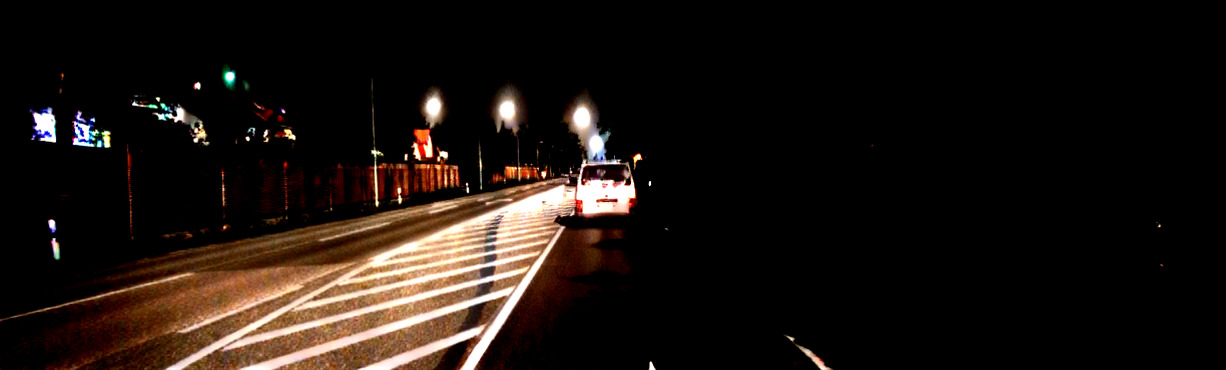}
    \caption{Day-to-Night}
\end{subfigure}
\hfill
\begin{subfigure}[b]{0.195\textwidth}
    \includegraphics[width=\textwidth]{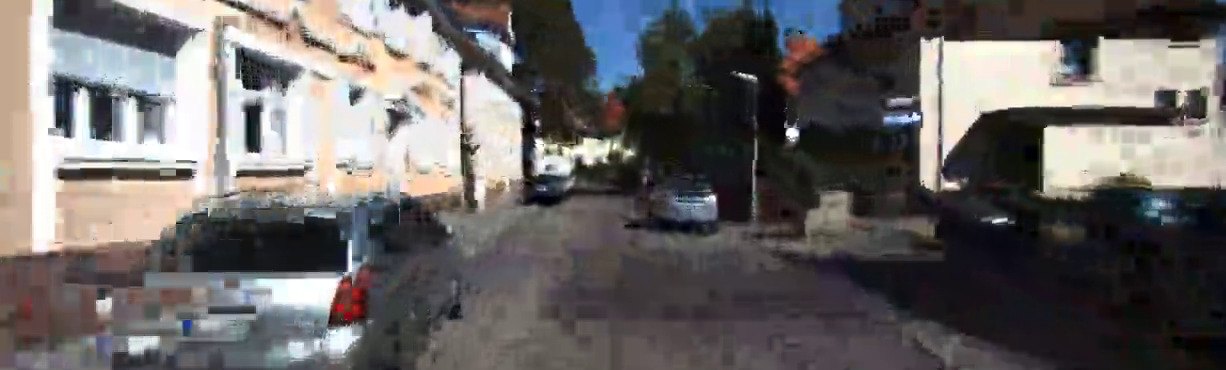}
    \caption{\mbox{Bandwidth (Severe)}}
\end{subfigure}
\\[2pt]
\begin{subfigure}[b]{0.195\textwidth}
    \includegraphics[width=\textwidth]{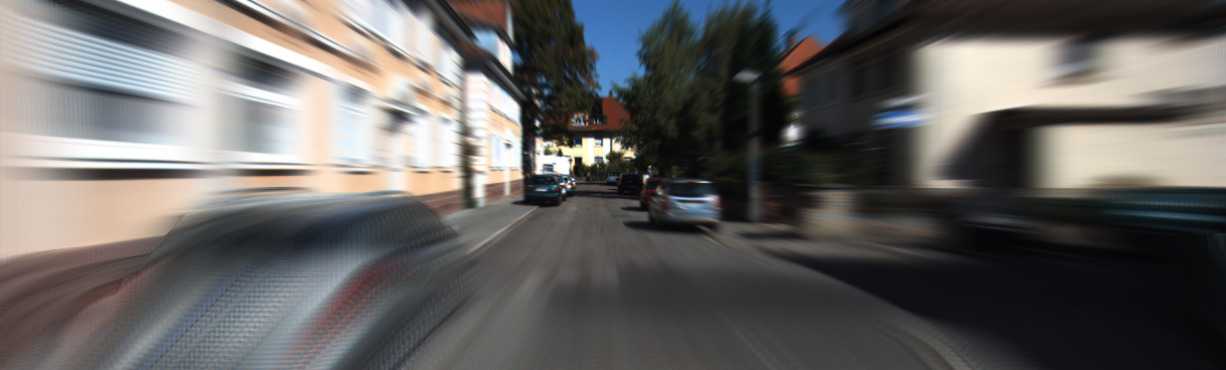}
    \caption{\mbox{Motion Blur (Severe)}}
\end{subfigure}
\hfill
\begin{subfigure}[b]{0.195\textwidth}
    \includegraphics[width=\textwidth]{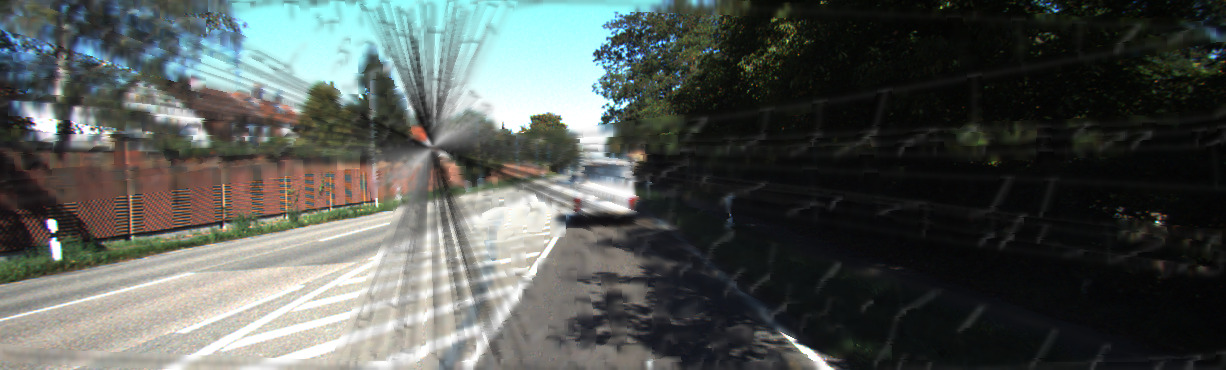}
    \caption{\mbox{Cracked Lens (Heavy)}}
\end{subfigure}
\hfill
\begin{subfigure}[b]{0.195\textwidth}
    \includegraphics[width=\textwidth]{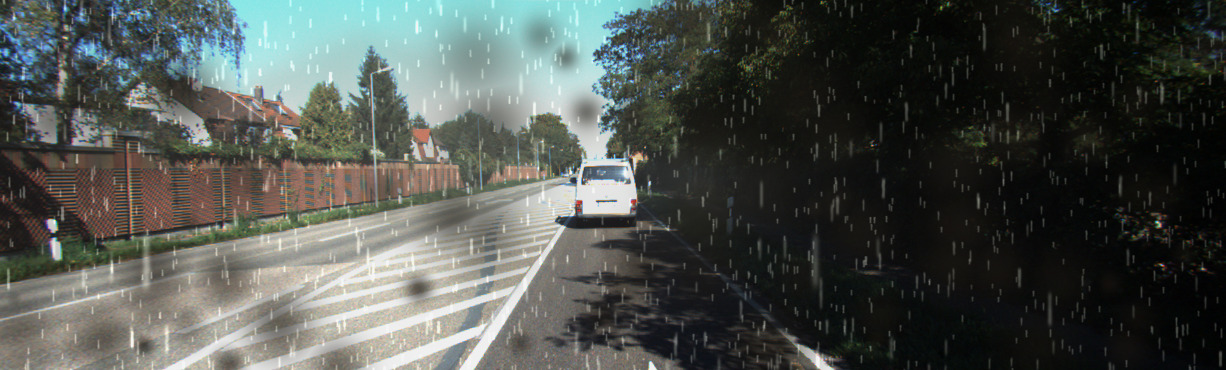}
    \caption{\mbox{Soiling+Rain (Heavy)}}
\end{subfigure}
\hfill
\begin{subfigure}[b]{0.195\textwidth}
    \includegraphics[width=\textwidth]{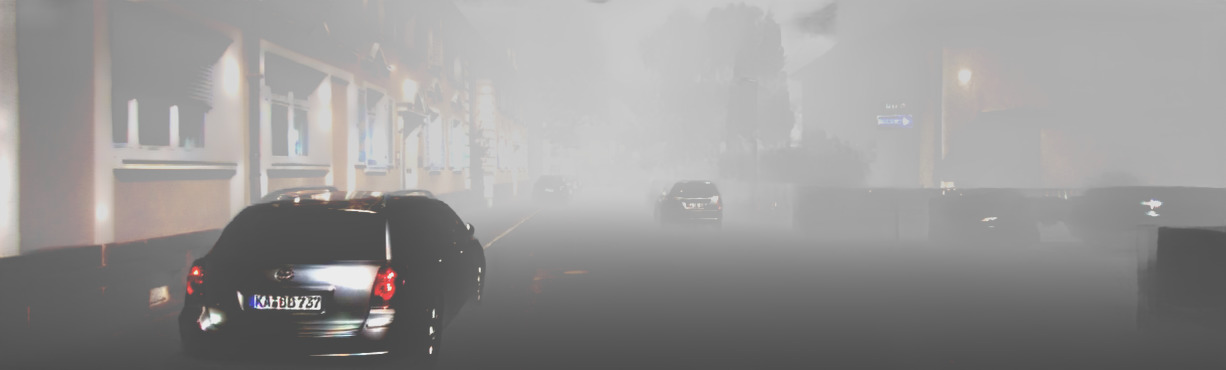}
    \caption{\mbox{Night+Fog (Moderate)}}
\end{subfigure}
\hfill
\begin{subfigure}[b]{0.195\textwidth}
    \includegraphics[width=\textwidth]{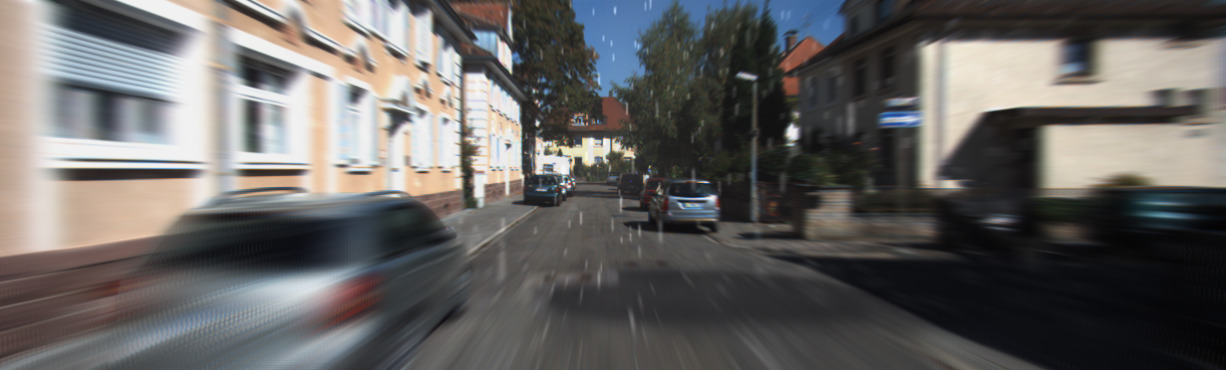}
    \caption{\mbox{Rain+Blur (Heavy)}}
\end{subfigure}
\\[1pt]
\caption{Example perturbations applied to KITTI sequences. Table~\ref{tab:exp_params} lists representative parameter values for the severity levels.}
\label{fig:perturbations}
\vspace{-15pt}
\end{figure*}

\textbf{Robustness boundary search.}
This pipeline reuses the decoupled perturbation and trajectory-evaluation
pipelines to estimate the perturbation severity at which a SLAM system
fails via the bisection procedure described in
Section~\ref{subsec:robustness_boundary}.
Users configure boundary evaluation by adding a \texttt{robustness\_boundary}
block (Listing~\ref{lst:boundary_yaml}) to the same experiment YAML
(Listing~\ref{lst:config}).
\texttt{target\_perturbation} selects which perturbation definition in the
\texttt{perturbations} list the boundary search uses. The search procedure (Section~\ref{subsubsec:search_procedure}) then
searches the \texttt{parameter} of the \texttt{target\_perturbation} between
\texttt{lower\_bound} and \texttt{upper\_bound}, while all other parameters remain fixed. It stops when the remaining
interval width (\texttt{upper\_bound} - \texttt{lower\_bound}) is at most
\texttt{tolerance} or when \texttt{max\_iters} is reached. The search marks a trial as
failure when ATE RMSE exceeds \texttt{ate\_rmse\_fail}.
\begin{lstlisting}[style=yaml, resetmargins=true, xleftmargin=0pt, xrightmargin=0pt, caption={Example robustness-boundary block.}, label=lst:boundary_yaml]
robustness_boundary:
  target_perturbation: fog_example
  parameter: visibility_m
  lower_bound: 10      # meters
  upper_bound: 200     # meters
  tolerance: 5         # stop when bound width <= 5 m
  max_iters: 10
  ate_rmse_fail: 1.5   # meters
\end{lstlisting}

\section{Perturbation Effects}
\label{sec:perturbations}

\noindent This section presents SAL's built-in perturbation modules, highlighting realistic condition modeling and the extensible perturbation-module interface.
We use existing techniques to implement these built-in perturbations as the purpose is
to showcase our extensibility.
We organize modules into weather and illumination (Section~\ref{subsec:scene_illumination}), camera (Section~\ref{subsec:camera}), and video transport (Section~\ref{subsec:video_transport}) conditions.
Figure~\ref{fig:perturbations} shows representative outputs of these built-in perturbation modules on KITTI sequences.

\subsection{Weather and Illumination Conditions}
\label{subsec:scene_illumination}

\noindent \textbf{Fog:} Fog reduces visibility and alters appearance,
making it a common challenge for outdoor autonomy. The fog module
implements the Koschmieder atmospheric scattering model~\cite{koschmieder},
which parameterizes fog severity by meteorological visibility distance
$V$ (meters) and uses it to set the extinction coefficient $\beta$ (1/m) via
$\beta = 3.912 / V$, making severity operationally meaningful.
Larger $\beta$
means stronger attenuation: the transmission $t(x)$ drops toward zero over
shorter distances, so distant regions appear more washed out than nearby ones.
We compute fog as follows~\cite{narasimhan2002vision,narasimhan2000chromatic}:
$I_{\text{fog}}(x) = J(x) \cdot t(x) + A \cdot (1 - t(x))$,
where $I_{\text{fog}}(x)$ is the foggy image, $J(x)$ is the clear scene radiance
at pixel $x$, $t(x) = \exp(-\beta d(x))$ is the transmission with depth $d(x)$
(meters), and $A$ is the atmospheric light. To simulate
spatially varying fog, the module supports heterogeneous fog by spatially
perturbing the extinction coefficient $\beta$ with procedural noise, yielding
non-uniform fog density. 

\begin{table*}[t]
\vspace*{1.2mm}
\centering
\caption{Experimental perturbation parameters. L, M, H, S denote light to severe levels. Top rows: single perturbations. Bottom: composite perturbations.}
\label{tab:exp_params}
\footnotesize
\setlength{\tabcolsep}{3pt}
\begin{tabular}{@{}llcccccr@{}}
\toprule
\textbf{Perturbation} & \textbf{Parameter} & \textbf{Light (L)} & \textbf{Moderate (M)}
& \textbf{Heavy (H)} & \textbf{Severe (S)} & \textbf{Dataset} \\
\midrule
Rain & Intensity (mm/h) & 3 & 10 & 50 & 200 & KITTI \\
Soiling & Particles (count) & 15 & 40 & 80 & 150 & KITTI, TUM \\
Cracked Lens & Force (N) / Threshold (N) & 300 / 400 & 400 / 350 & 500 / 300 &
600 / 250 & TUM \\
Bandwidth Compression & Target / Max / Buffer (Mbps) & 5 / 5 / 10 & 3 / 3 / 6 & 1 / 1 / 2 & 0.3 / 0.3 / 0.6 & EuRoC \\
Frame Drop & Drop Rate (\%) (random) & 10 & 20 & 30 & 50 & EuRoC \\
\midrule
Rain + Soiling & Intensity (mm/h) / Particles (count) & 3 / 15 & 10 / 40 & 50 /
80 & 200 / 150 & KITTI \\
Night + Fog & Visibility (m) & 200 & 50 & 20 & 10 & KITTI \\
Rain + Motion Blur & Int. (mm/h) / Speed (km/h) / Exposure (ms) & 3 / 30 / 33 & 10 /
50 / 50 & 50 / 80 / 66 & 200 / 120 / 100 & KITTI \\
Soiling + Crack & Particles / Force / Threshold (N) & 15 / 300 / 400 & 40 / 400
/ 350 & 80 / 500 / 300 & 150 / 600 / 250 & TUM \\
\bottomrule
\end{tabular}
\vspace{-15pt}
\end{table*}

\textbf{Rain:} Rain reduces visibility and occludes camera views in outdoor
operation. The rain module uses a physics-based particle rendering
pipeline~\cite{tremblay2020rain,de2012fast}, parameterized by rainfall rate (mm/h), which
makes severity meteorologically meaningful and reports robustness by rain
intensity. It models haze-like attenuation from
distant sub-pixel droplets and dynamic streak
occlusions from nearby visible drops. Given an RGB image and depth map, it
first generates a fog-like attenuation layer for distant droplets, then uses a
particle simulator~\cite{de2012fast} to compute streak positions from camera
calibration. Streaks are drawn with templates from the Garg and Nayar
rain-streak database~\cite{garg2007rain}, with radiance scaled from an
environment map estimated from the input image, and composited with
depth-of-field blur.

\textbf{Day-to-Night:} Autonomous systems often operate across varying lighting
conditions, yet most datasets are captured in daytime. This module
transforms daytime scenes to nighttime using
CycleGAN-Turbo~\cite{zhu2017cyclegan,isola2017pix2pix}, a one-step image
translation model applied frame-wise. Since translation is per-frame, some temporal flicker may occur, but the model
largely preserves spatial layout, making it suitable for SLAM evaluation and
demonstrating the framework's ability to support both learned and model-based
perturbation modules.

\subsection{Camera Conditions}
\label{subsec:camera}

\noindent \textbf{Lens Soiling:} Robots encounter mud splashes, dust accumulation,
and water droplets on camera lenses. The soiling module renders lens particles
as dark bokeh spots that partially occlude the scene, with configurable size
(pixel radius), opacity (light absorption), and color. Each particle uses
a Gaussian falloff (dark center, soft edges), and overlapping particles use max
blending (higher opacity wins) to avoid unrealistically black regions. 
The contamination pattern is held constant across frames to preserve temporal consistency.

\textbf{Cracked Lens:} This module uses an open-source
implementation~\cite{prabhakar2024fractured} to simulate glass fracture on a graph of
randomly sampled points using physically motivated parameters (impact force and
break threshold, in N). An impact point and force set the initial stress. Stress
propagates to nearby points within a radius, and edges whose stress exceeds a
break threshold become crack segments. The algorithm forms secondary crack lines by connecting stressed points with a minimum spanning tree and renders the resulting pattern. The module generates the crack texture once per sequence and applies a localized Gaussian blur around crack regions to mimic out-of-focus fracture edges.

\textbf{Motion Blur:} Fast camera motion causes blur that degrades feature detection. Rapid forward camera translation produces a radial blur
field expanding from the focus of expansion (FoE), located near the principal
point $(c_x,c_y)$~\cite{hartley2003multiple}: pixels farther from the FoE and closer in depth have larger
displacements during the exposure. Our module synthesizes this effect from
camera speed $s$ (km/h) and exposure time $e$ (ms), tying blur severity
directly to measurable motion and camera timing, by backprojecting each pixel
using per-pixel depth, reprojecting after a forward translation $\Delta Z =
(s/3.6)\,(e/1000)$ (meters), and blurring along the resulting per-pixel
image-plane motion vector.
We assume pure forward translation, a static scene, and a global-shutter camera.
This isolates depth-dependent motion blur as a controlled experimental variable. Real
driving can also involve rotation and independently moving objects, which this
forward-motion blur module does not explicitly model.

\subsection{Video Transport Conditions}
\label{subsec:video_transport}

\noindent \textbf{Bandwidth Compression:} Wireless or cloud-connected robots often process bandwidth-limited video streams. This
module re-encodes each sequence with H.264~\cite{wiegand2003h264} via FFmpeg
(libx264, codec configurable) under constant-bitrate (CBR) control
using a target bitrate, a max bitrate cap, and a VBV buffer size. The maxrate is
the highest rate allowed into the decoder buffer, and the VBV buffer size is the
buffer capacity that permits short-term rate variation around the target. This
models bandwidth-constrained streaming where the encoder must keep to a fixed
bitrate. When scene complexity exceeds that budget, compression becomes more
aggressive, producing blocking and temporal smearing.

\textbf{Frame Drop:} Frame drops arise from network packet loss, sensor
glitches, or bandwidth constraints. The module supports random frame dropping
with a configurable probability (e.g., 30\%). The module also supports periodic dropping of every
$N$-th frame. It removes dropped frames from the
sequence and the corresponding metadata entries (e.g., for EuRoC~\cite{burri2016euroc}, timestamp-image rows in \texttt{cam0\_data.csv}) to create realistic gaps.

\section{Evaluation}
\label{sec:experiments}

\noindent We demonstrate SAL’s broad coverage and extensibility across perturbation modules, SLAM algorithms, and datasets.
To evaluate SAL's capabilities, we apply perturbations across seven SLAM systems and three datasets, measure trajectory accuracy, analyze feature-tracking behavior, and run robustness-boundary search on configurations
that exhibit tracking failure.

\subsection{Experimental Setup}
\label{subsec:setup}

\noindent All experiments ran on a workstation with a 12-core Intel Core i7-12700K CPU
(at 4.9\,GHz), an NVIDIA RTX 3090 GPU (10496 CUDA cores, at 1.9\,GHz), and 64\,GB
RAM.

Table~\ref{tab:exp_params} lists the perturbations and severity levels used in our experiments. The listed values are the specific settings used for our evaluation, users can configure different values in their experiment YAML.
For perturbation modules, we evaluate weather, camera, and transport
degradations, in both single and composite settings.
For datasets, we use KITTI~\cite{Geiger2012CVPR}, TUM~\cite{sturm12iros}, and EuRoC~\cite{burri2016euroc} to cover outdoor and indoor scenes
with common visual SLAM sensing modes (monocular, stereo, and RGB-D). For SLAM algorithms,
we evaluate ORB-SLAM3~\cite{Campos2021ORBSLAM3}, S3PO-GS~\cite{cheng2025s3pogs},
GigaSLAM~\cite{gigaslam2025}, MASt3R-SLAM~\cite{murai2024mast3rslam},
DROID-SLAM~\cite{teed2021droid}, VGGT-SLAM~\cite{maggio2025vggt-slam}, and
Photo-SLAM~\cite{hhuang2024photoslam} to cover classical feature-based and
learned/neural SLAM methods.
For feature-tracking analysis, we use ORB~\cite{murORB2} and
SuperPoint~\cite{detone2018superpoint} to cover classical and learned features.
For robustness-boundary evaluation, we focus on configurations that failed
during severity sweeps, and run boundary search with
\texttt{max\_iters}=10 and \texttt{ate\_rmse\_fail}=1.5\,m. The tolerance is
5\,m for night+fog visibility and 3 percentage points for frame drop.
Tables~\ref{tab:results}--\ref{tab:boundary_results} report 3-run averages.

\begin{figure*}[!t]
\vspace*{1mm}
\centering
\setlength{\tabcolsep}{1pt}
\begin{tabular}{@{}c ccccc@{}} & \scriptsize\textbf{Rain} & \scriptsize\textbf{Soiling} &
\scriptsize\textbf{Rain+Soil.} & \scriptsize\textbf{Night+Fog} & \scriptsize\textbf{Rain+Blur}
\\[1pt]
\rotatebox{90}{\parbox{1.2cm}{\centering\scriptsize\textbf{ORBSLAM3}}} &
\includegraphics[width=0.18\textwidth,height=1.6cm,keepaspectratio]{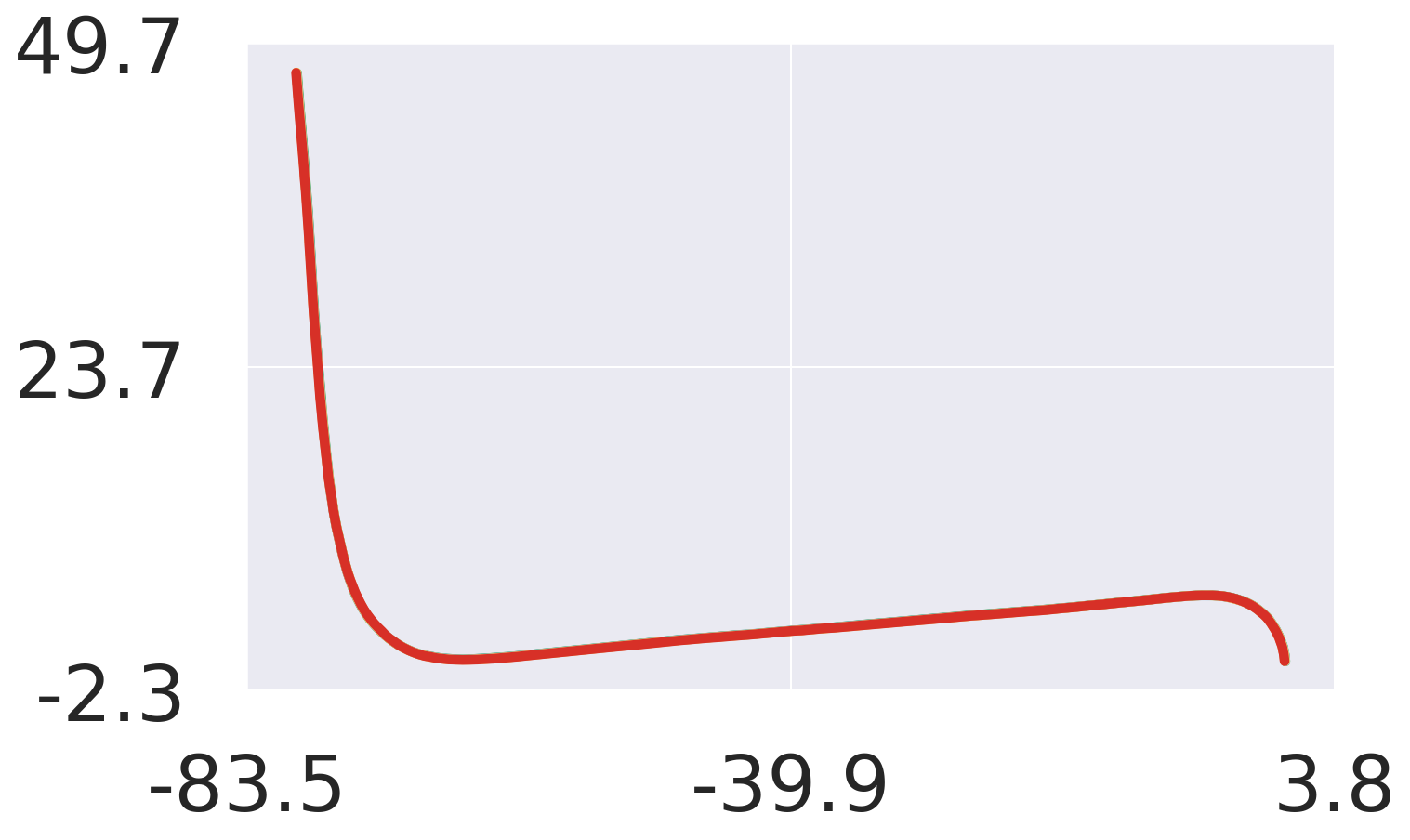} &
\includegraphics[width=0.18\textwidth,height=1.6cm,keepaspectratio]{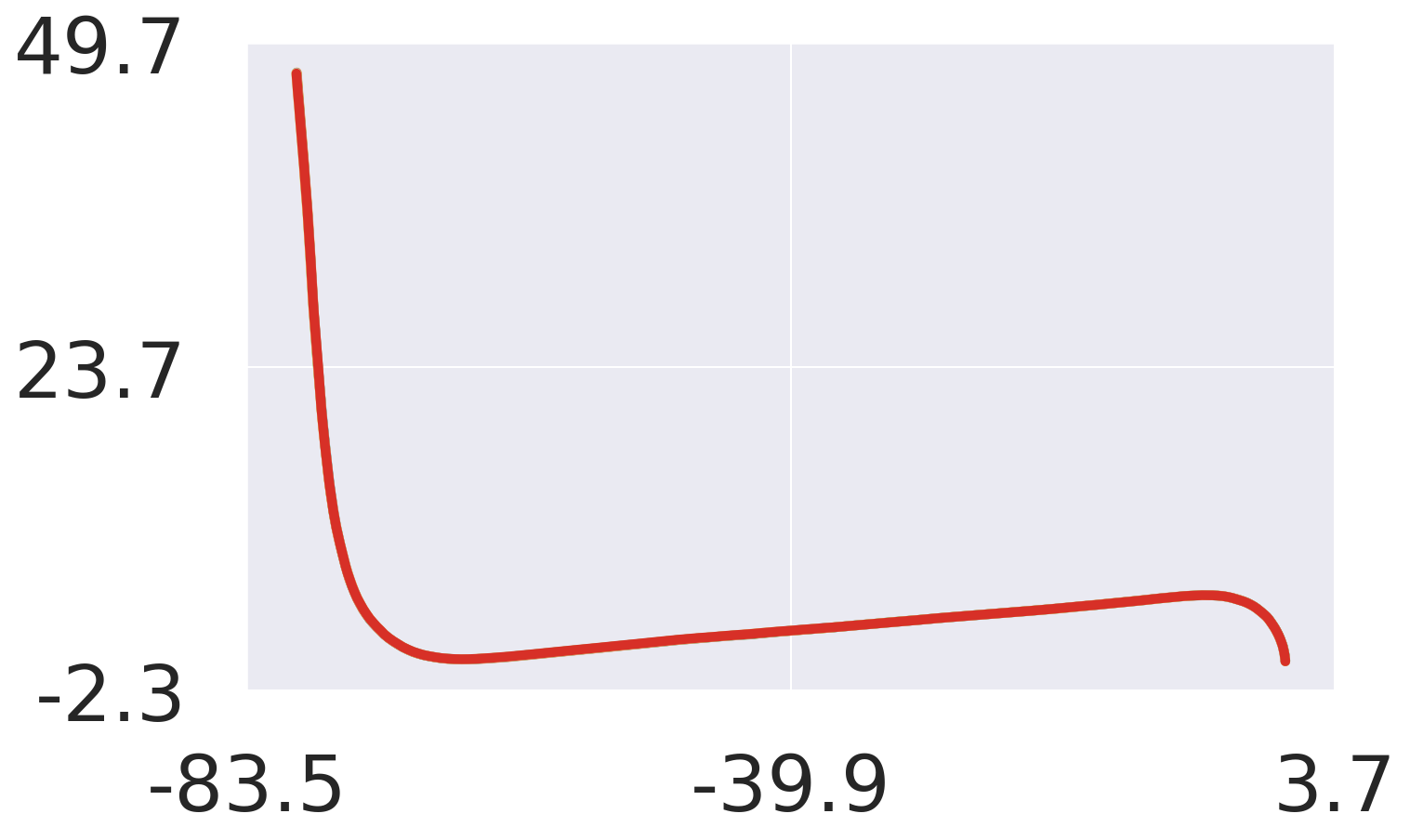} &
\includegraphics[width=0.18\textwidth,height=1.6cm,keepaspectratio]{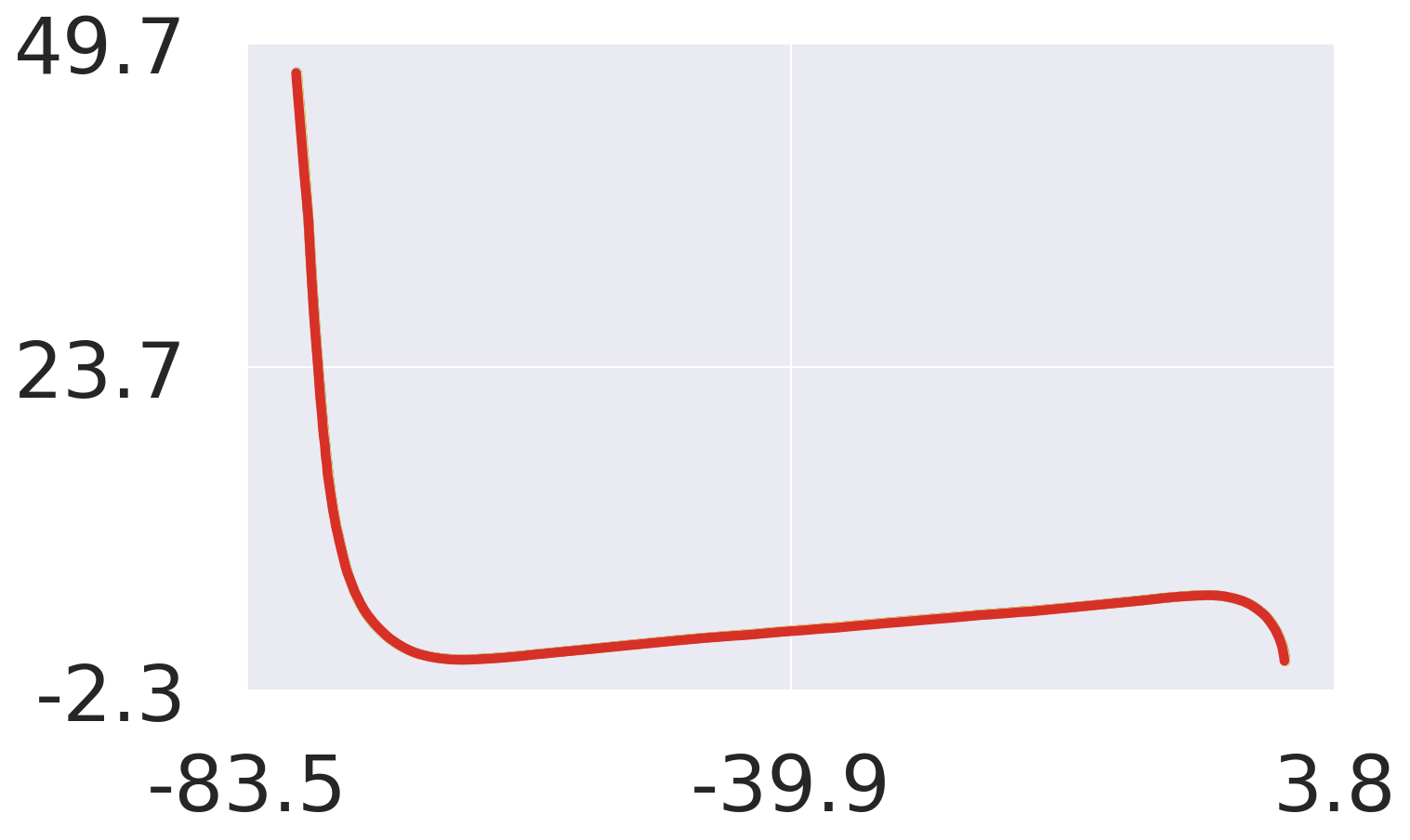} &
\includegraphics[width=0.18\textwidth,height=1.6cm,keepaspectratio]{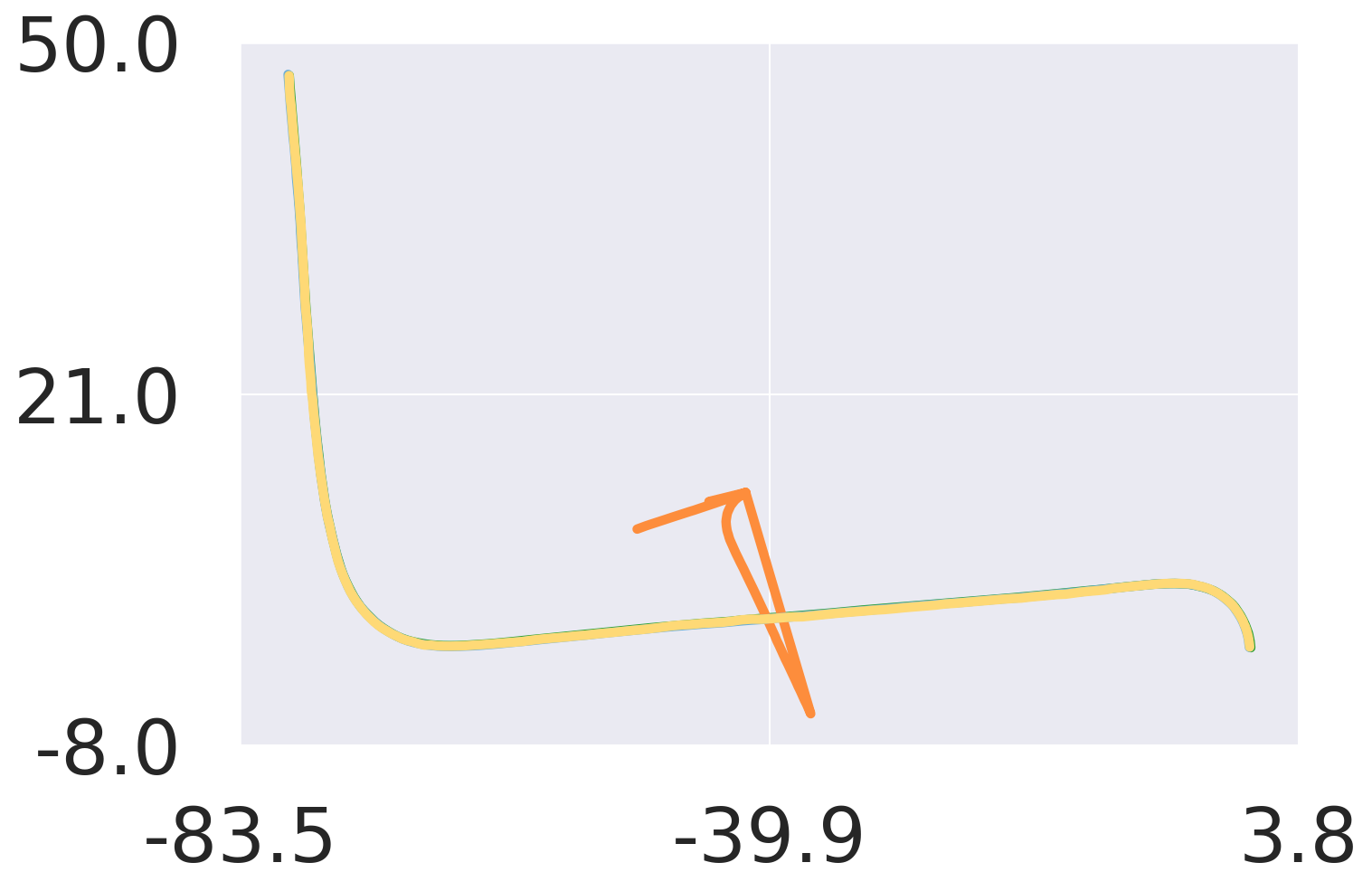} &
\includegraphics[width=0.18\textwidth,height=1.6cm,keepaspectratio]{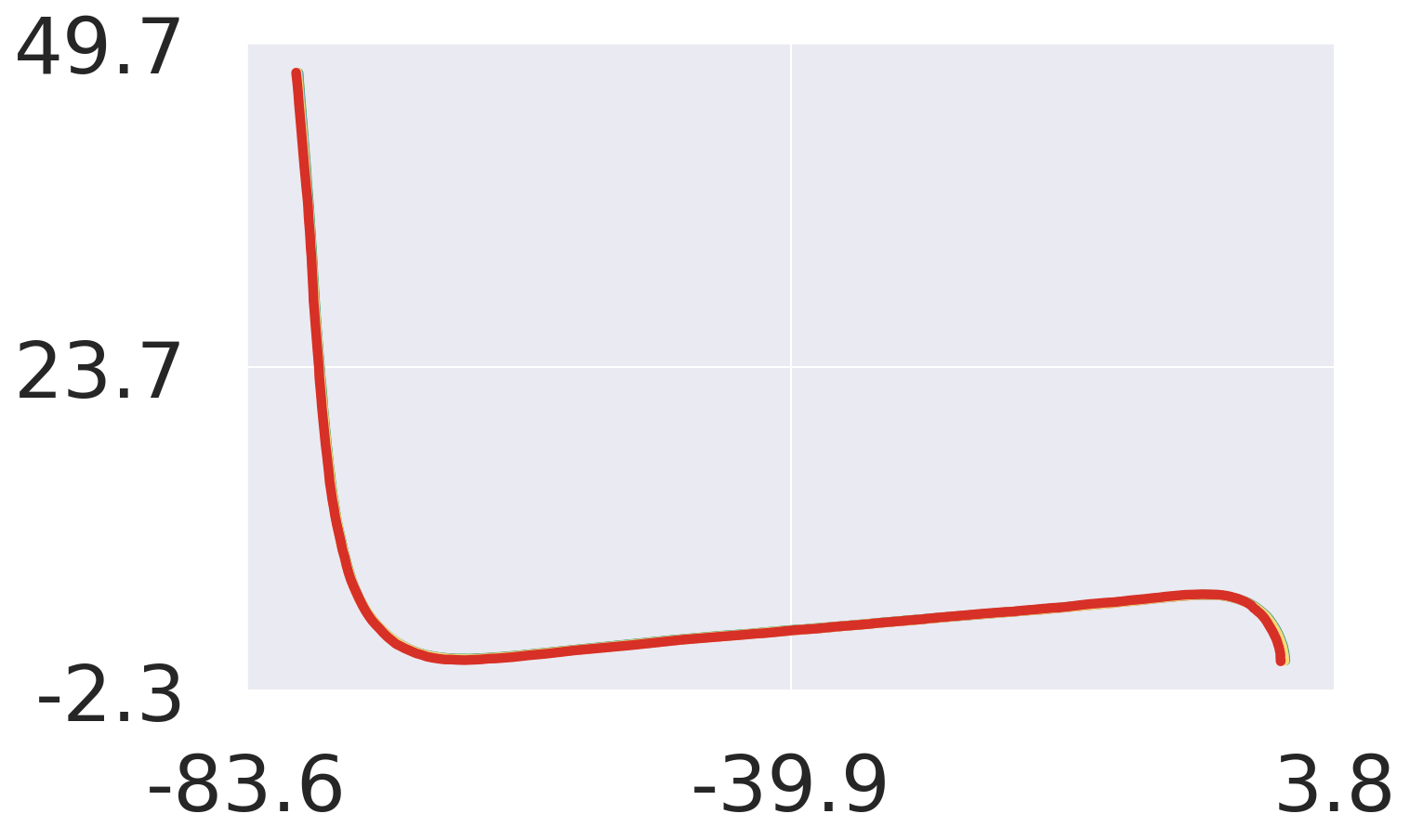}
\\[1pt]
\rotatebox{90}{\parbox{1.2cm}{\centering\scriptsize\textbf{S3PO-GS}}} &
\includegraphics[width=0.18\textwidth,height=1.6cm,keepaspectratio]{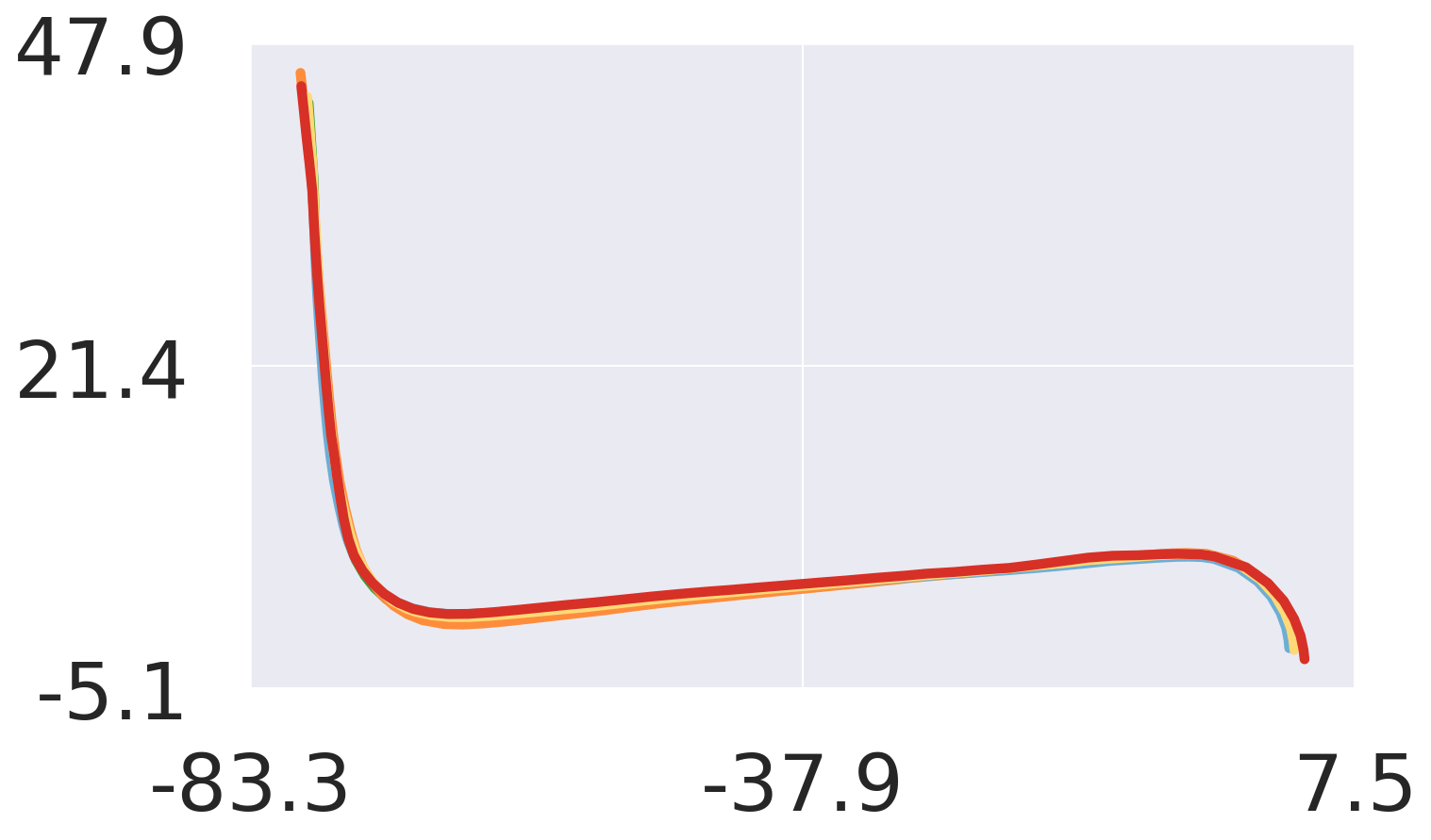} &
\includegraphics[width=0.18\textwidth,height=1.6cm,keepaspectratio]{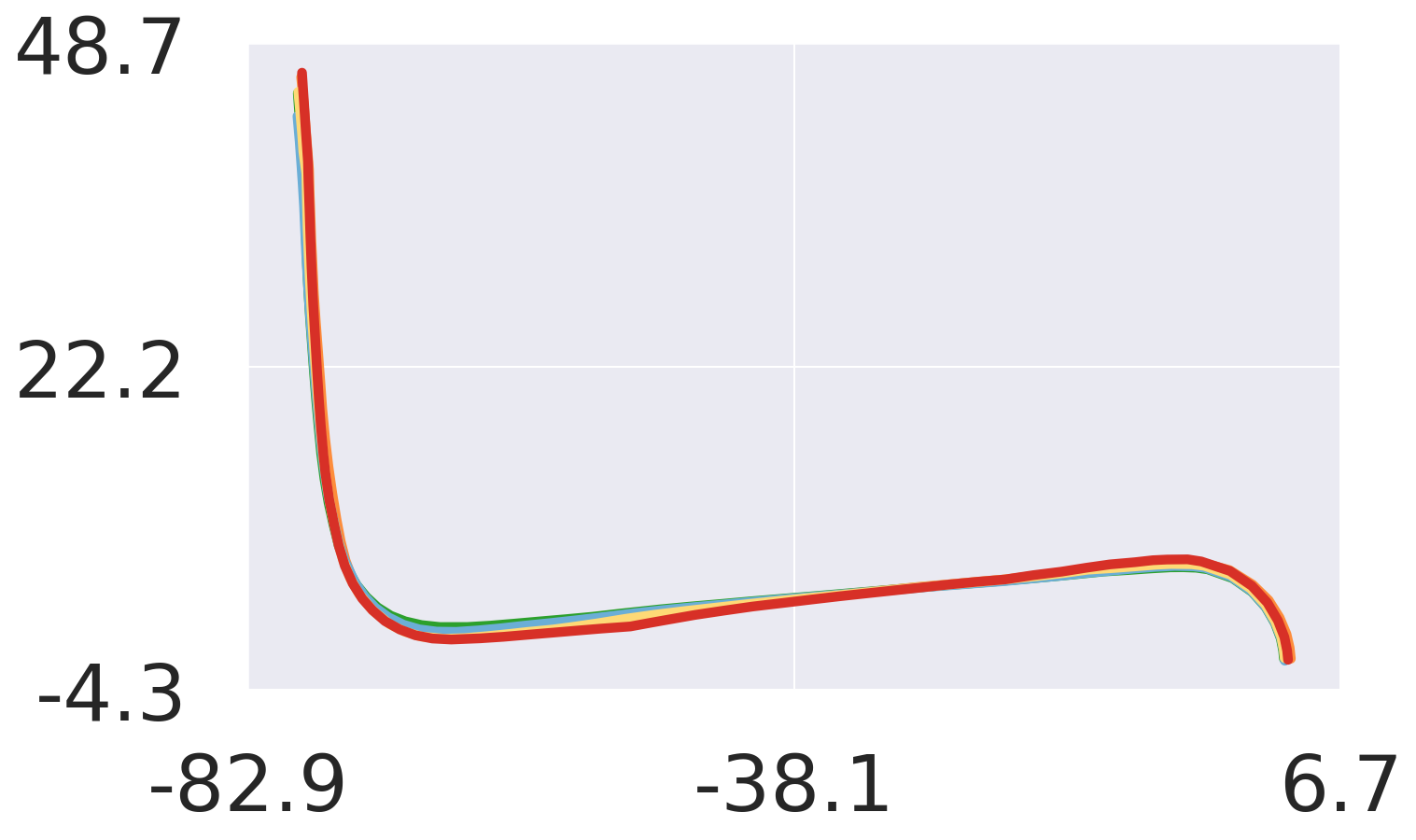} &
\includegraphics[width=0.18\textwidth,height=1.6cm,keepaspectratio]{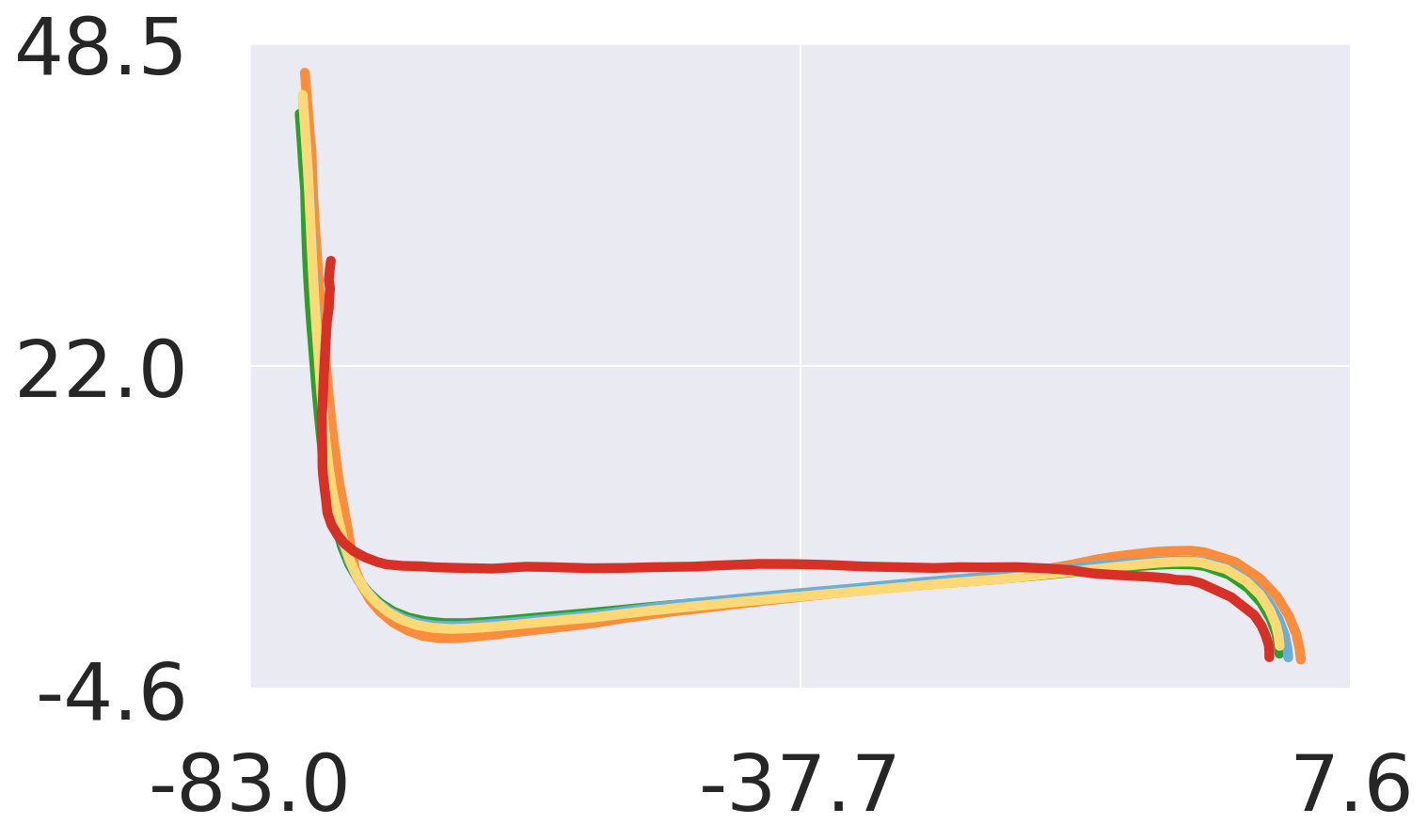} &
\includegraphics[width=0.18\textwidth,height=1.6cm,keepaspectratio]{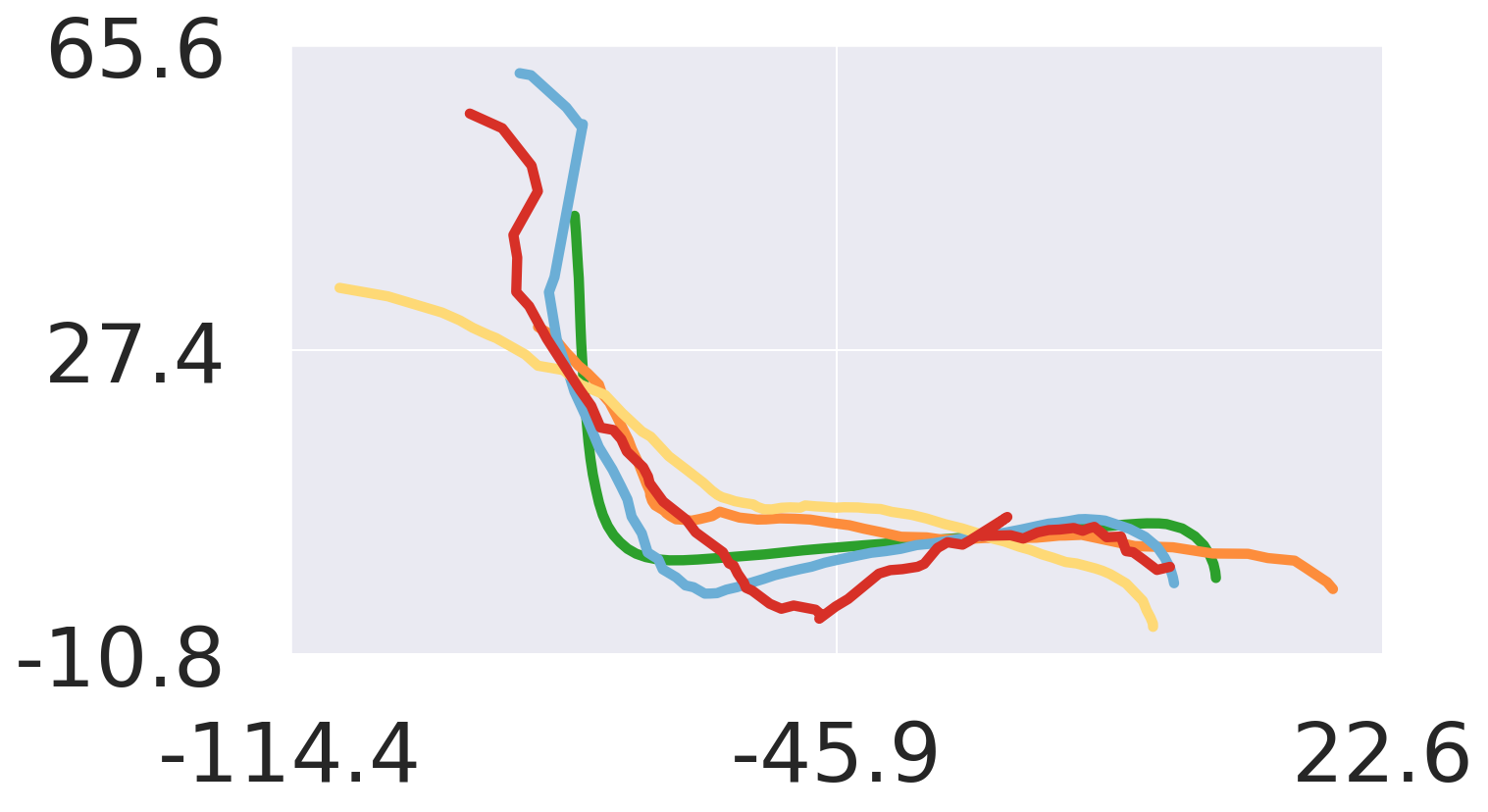} &
\includegraphics[width=0.18\textwidth,height=1.6cm,keepaspectratio]{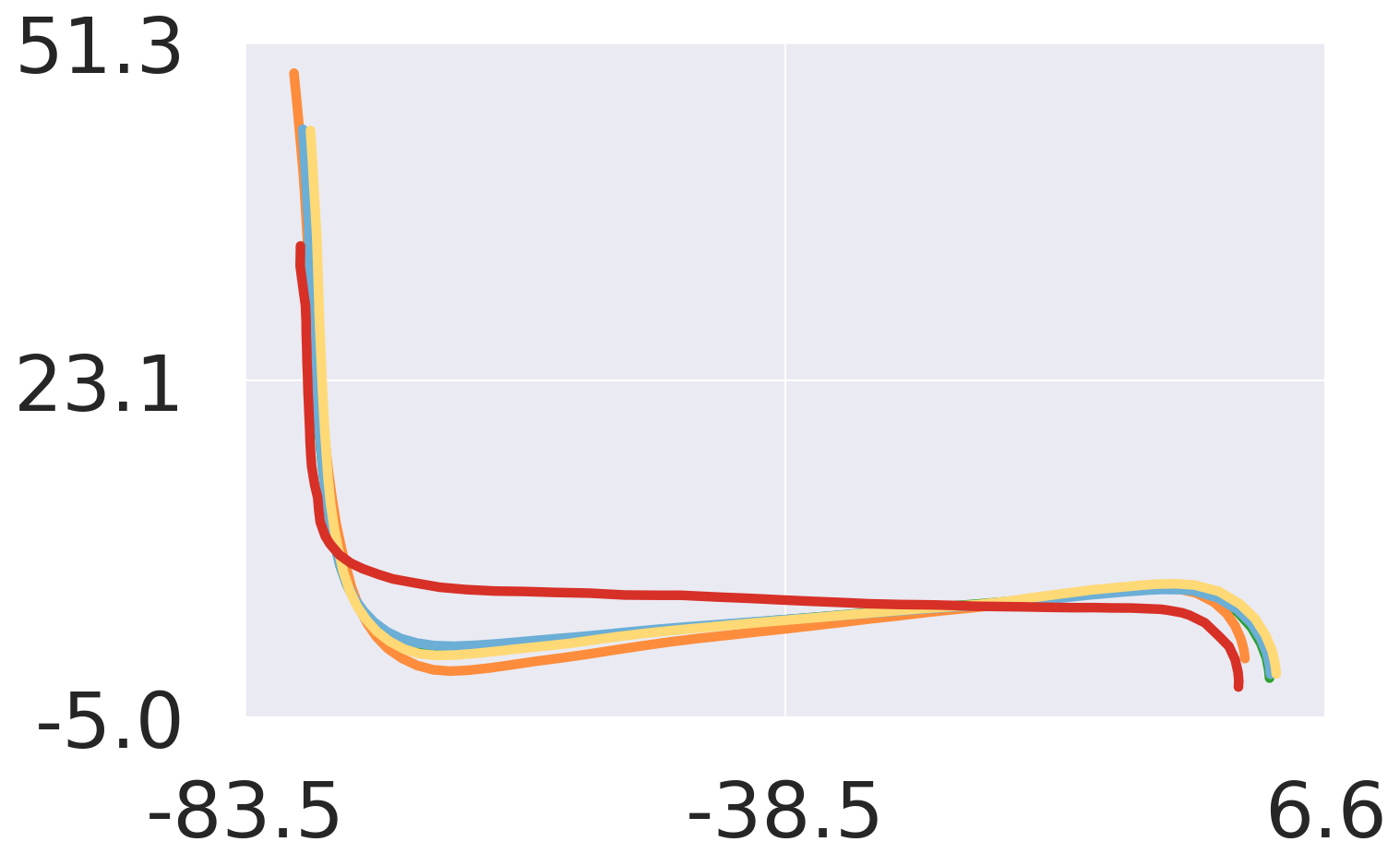}
\\[1pt]
\rotatebox{90}{\parbox{1.2cm}{\centering\scriptsize\textbf{GigaSLAM}}} &
\includegraphics[width=0.18\textwidth,height=1.6cm,keepaspectratio]{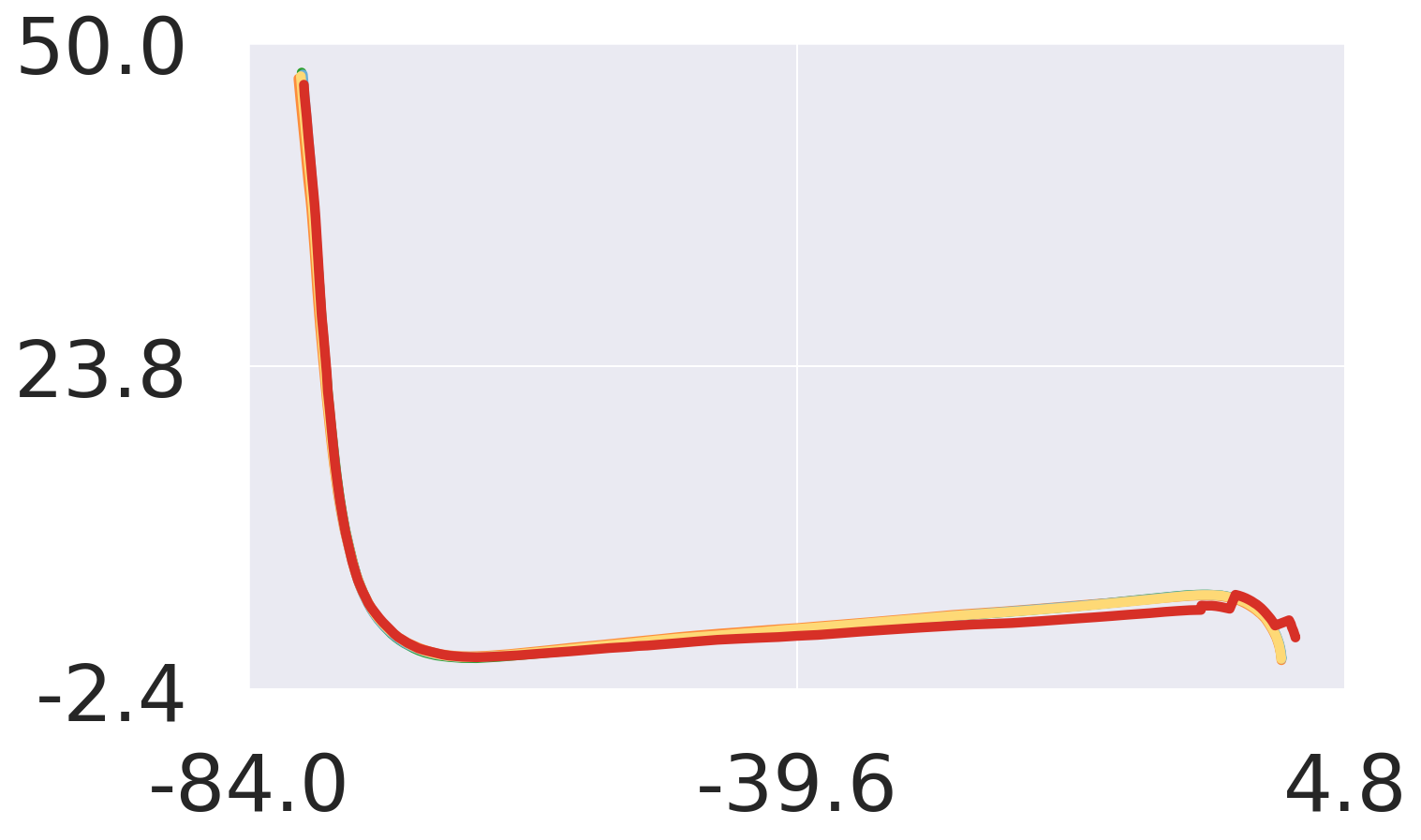} &
\includegraphics[width=0.18\textwidth,height=1.6cm,keepaspectratio]{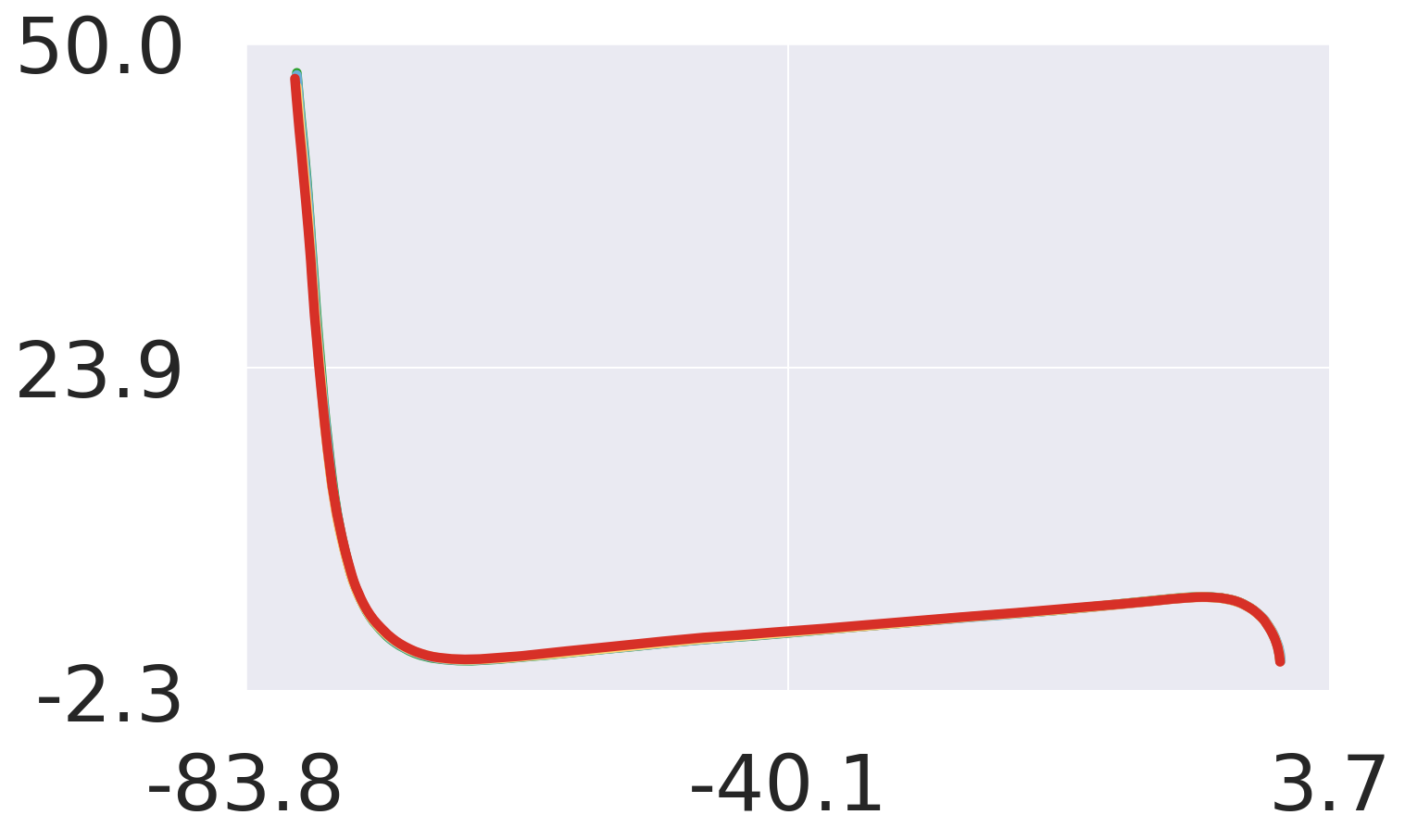} &
\includegraphics[width=0.18\textwidth,height=1.6cm,keepaspectratio]{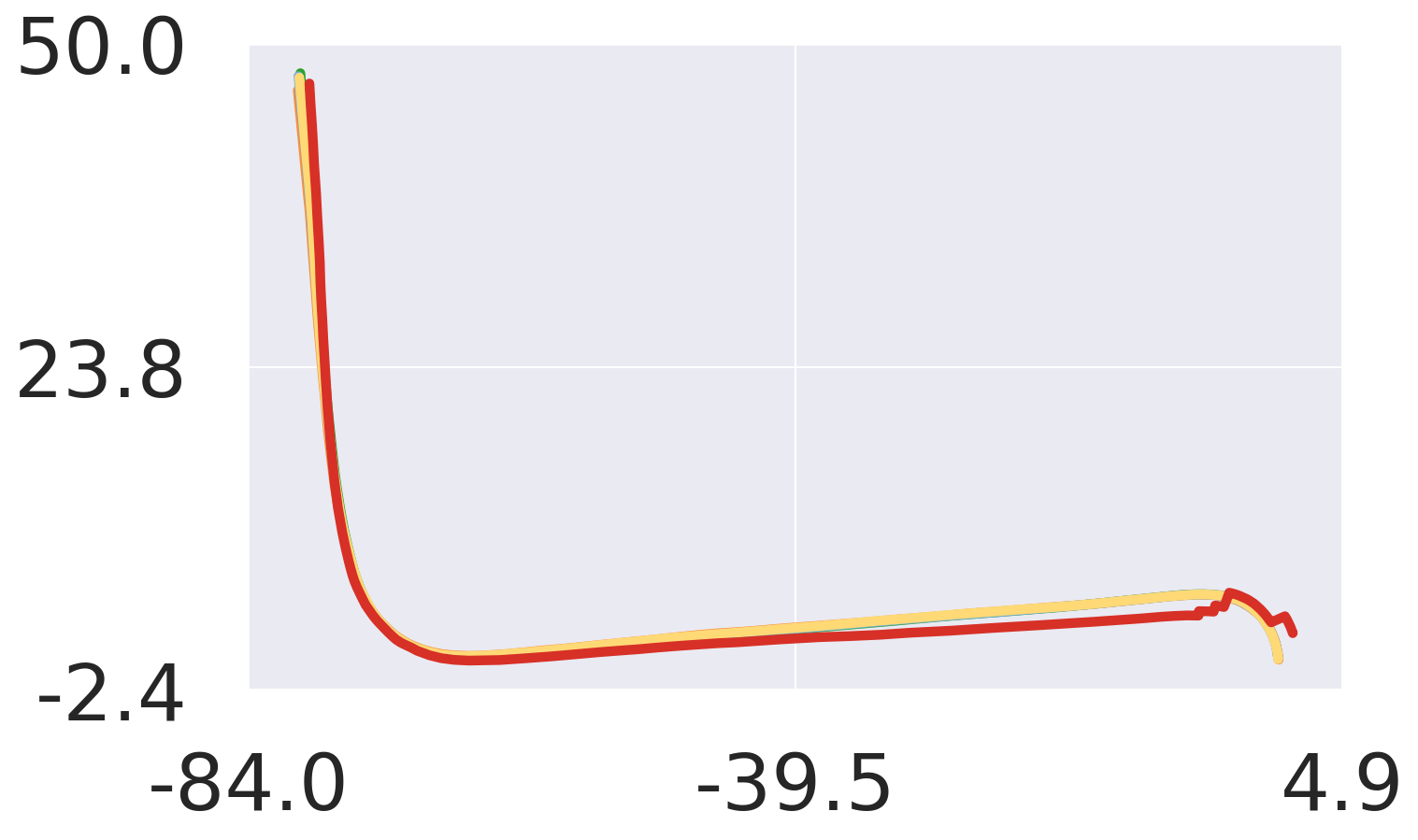} &
\includegraphics[width=0.18\textwidth,height=1.6cm,keepaspectratio]{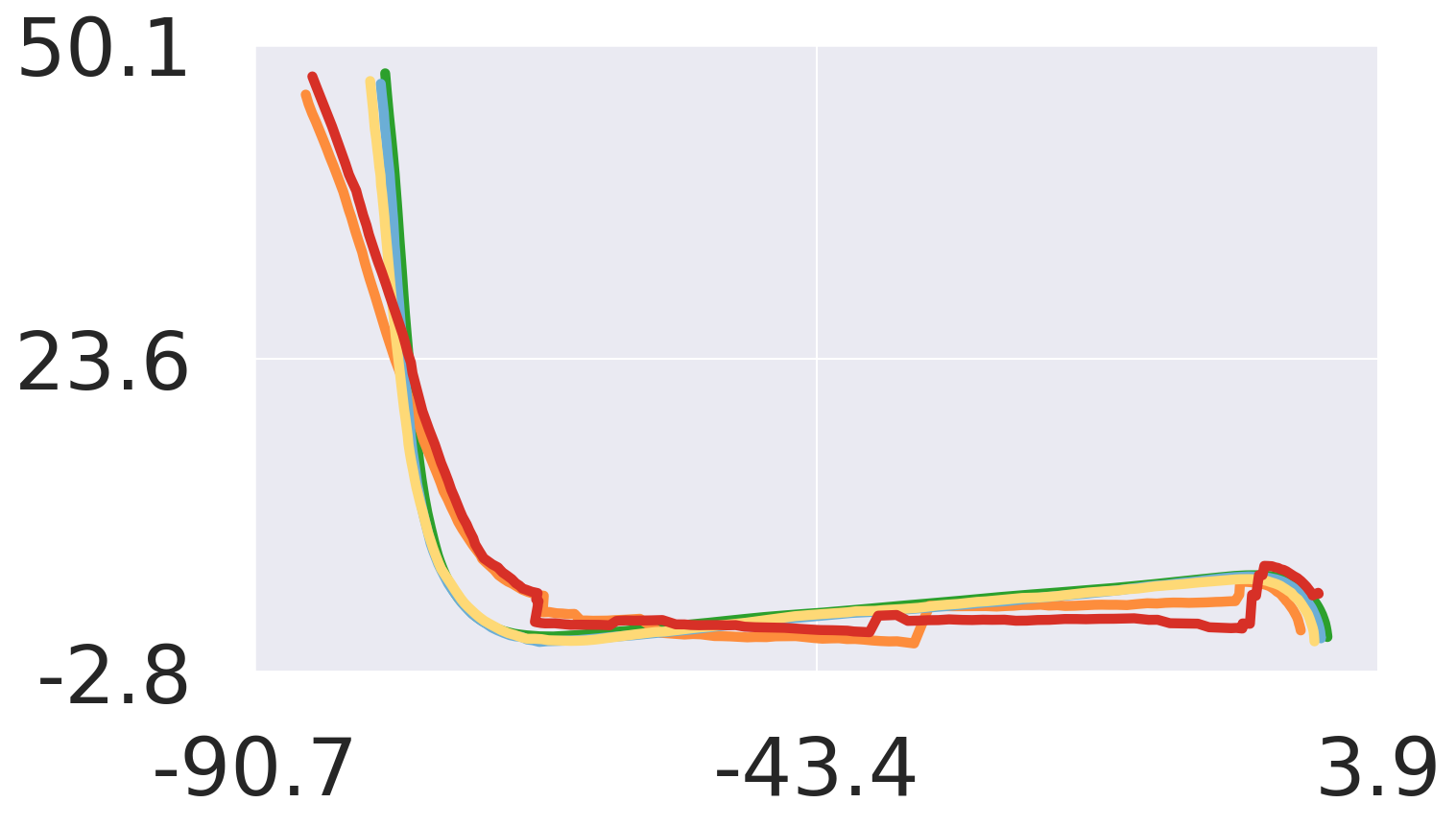} &
\includegraphics[width=0.18\textwidth,height=1.6cm,keepaspectratio]{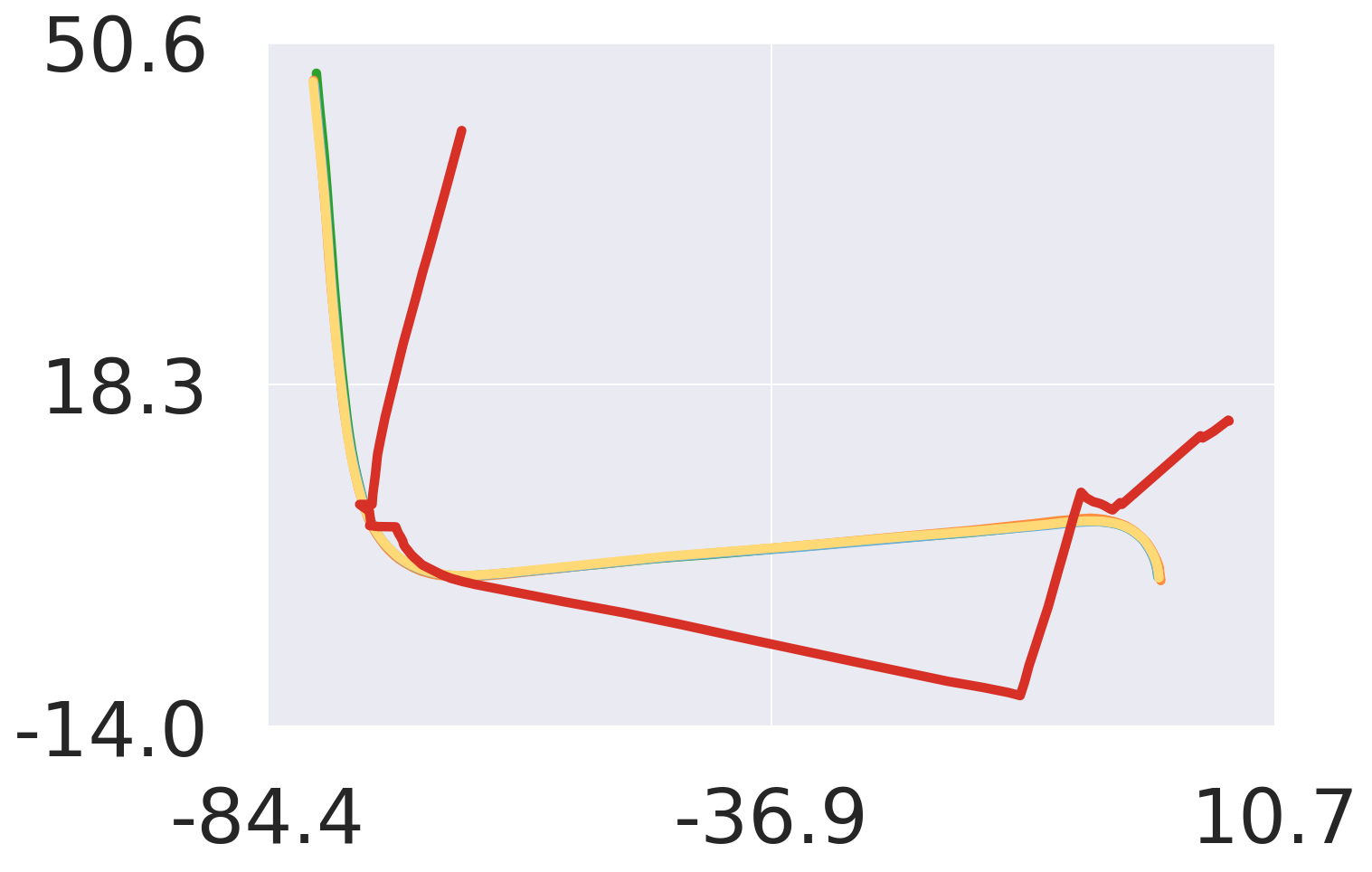} \\
\end{tabular}
\caption{Outdoor (KITTI) SLAM trajectory evaluation in the X-Z plane.
\legendswatch{2ca02c} baseline,
\legendswatch{6baed6} Light,
\legendswatch{fed976} Moderate,
\legendswatch{fd8d3c} Heavy,
\legendswatch{d73027} Severe.}
\label{fig:slam_results_kitti}
\vspace{-10pt}
\end{figure*}

\begin{figure}[!t]
\centering
\setlength{\tabcolsep}{0.5pt}
\begin{minipage}[t]{0.55\columnwidth}
\centering
\begin{tabular}{@{}c ccc@{}}
 & \scriptsize\textbf{Soiling} & \scriptsize\textbf{Crack} & \scriptsize\textbf{Soil.+Crack} \\[0pt]
\rotatebox{90}{\parbox{1.2cm}{\centering\scriptsize\textbf{ORB3}}} &
\includegraphics[width=0.30\columnwidth]{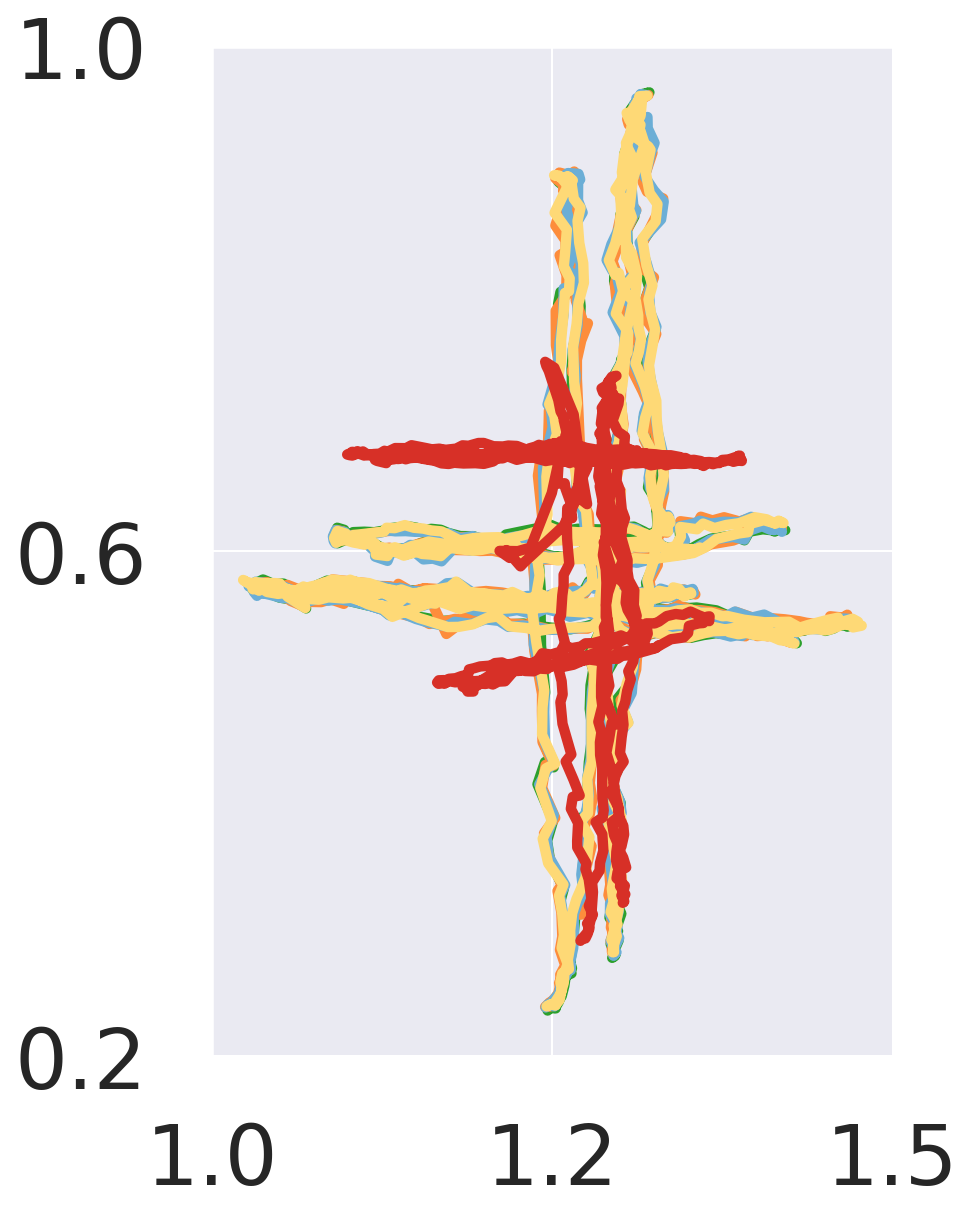} &
\includegraphics[width=0.30\columnwidth]{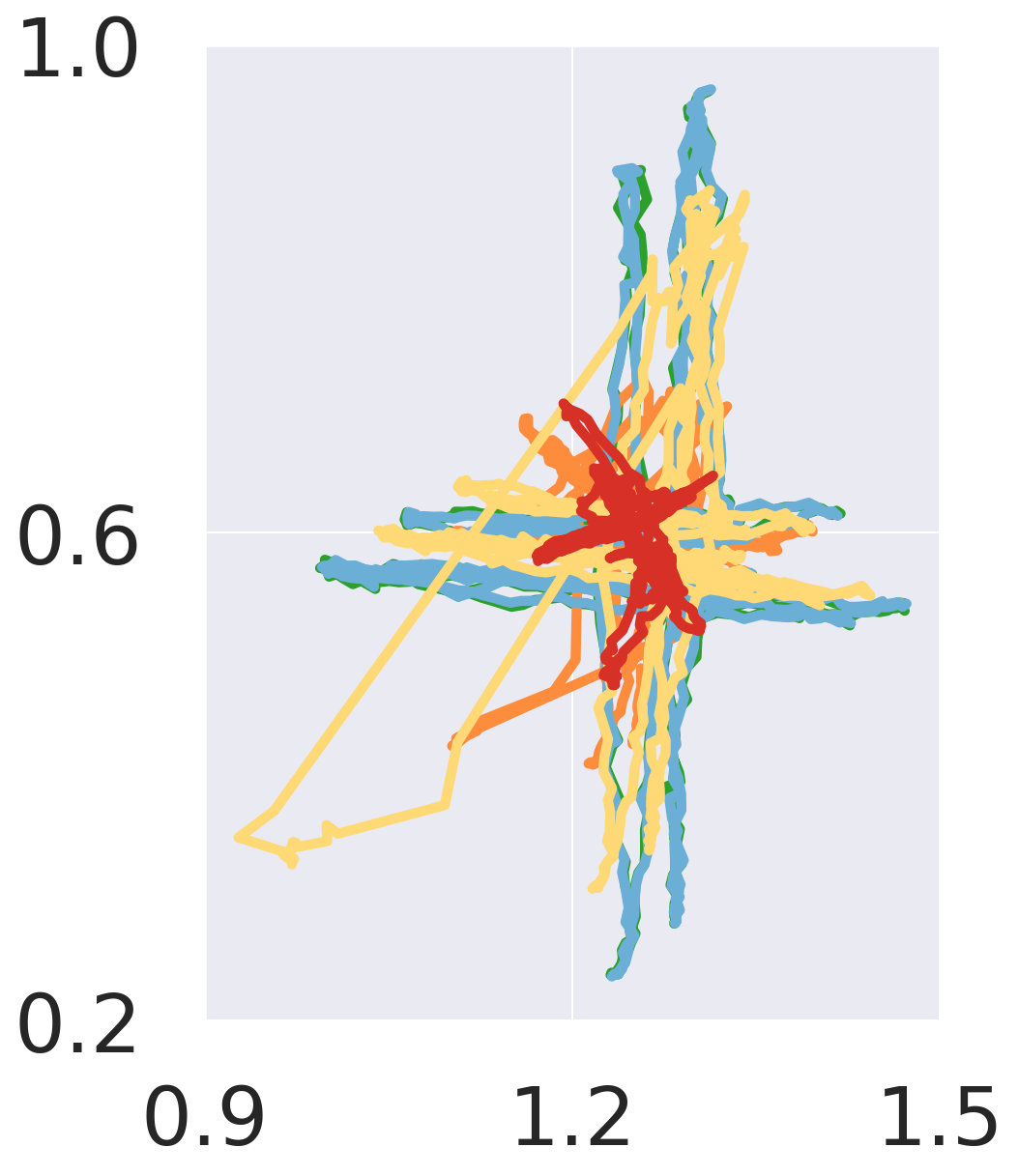} &
\includegraphics[width=0.30\columnwidth]{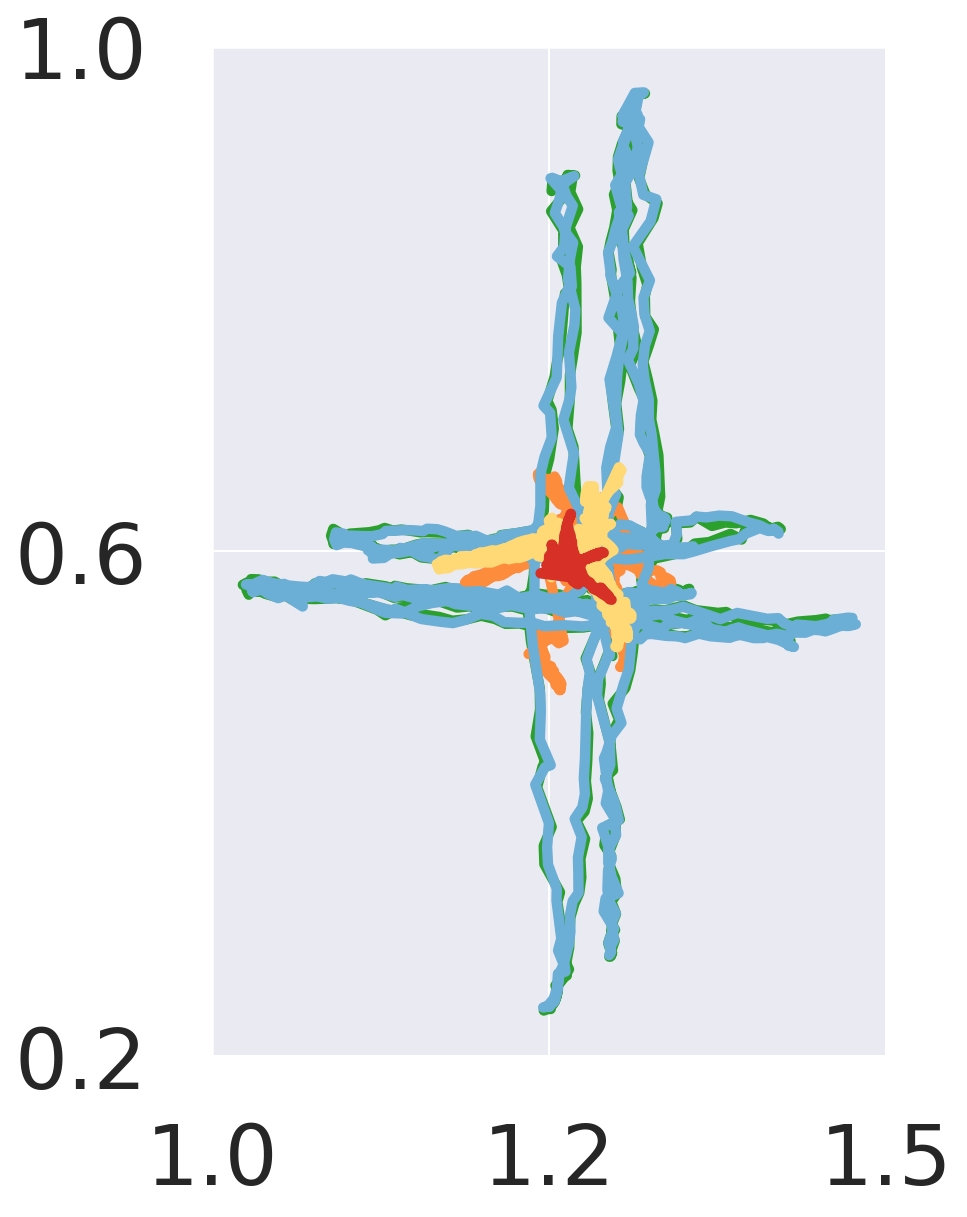}
\\[0pt]
\rotatebox{90}{\parbox{1.2cm}{\centering\scriptsize\textbf{MASt3R}}} &
\includegraphics[width=0.30\columnwidth]{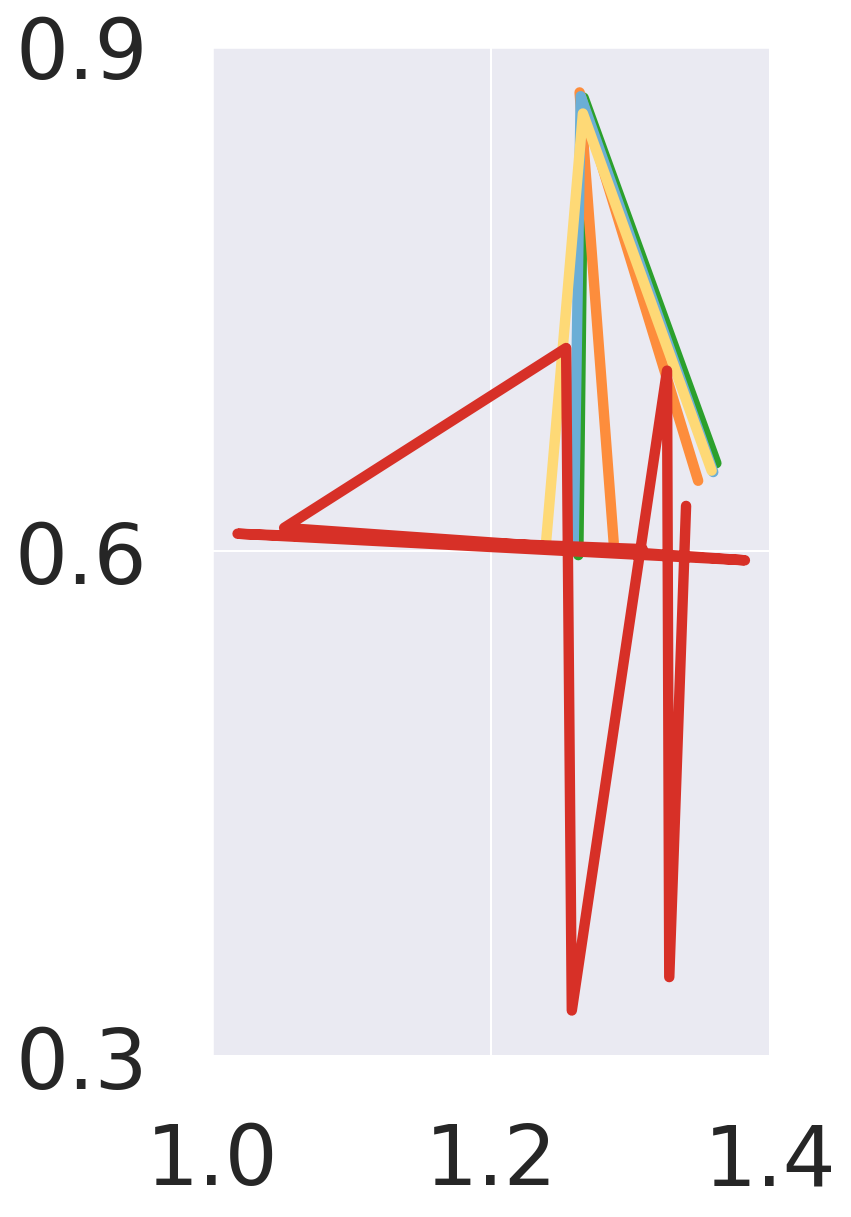} &
\includegraphics[width=0.30\columnwidth]{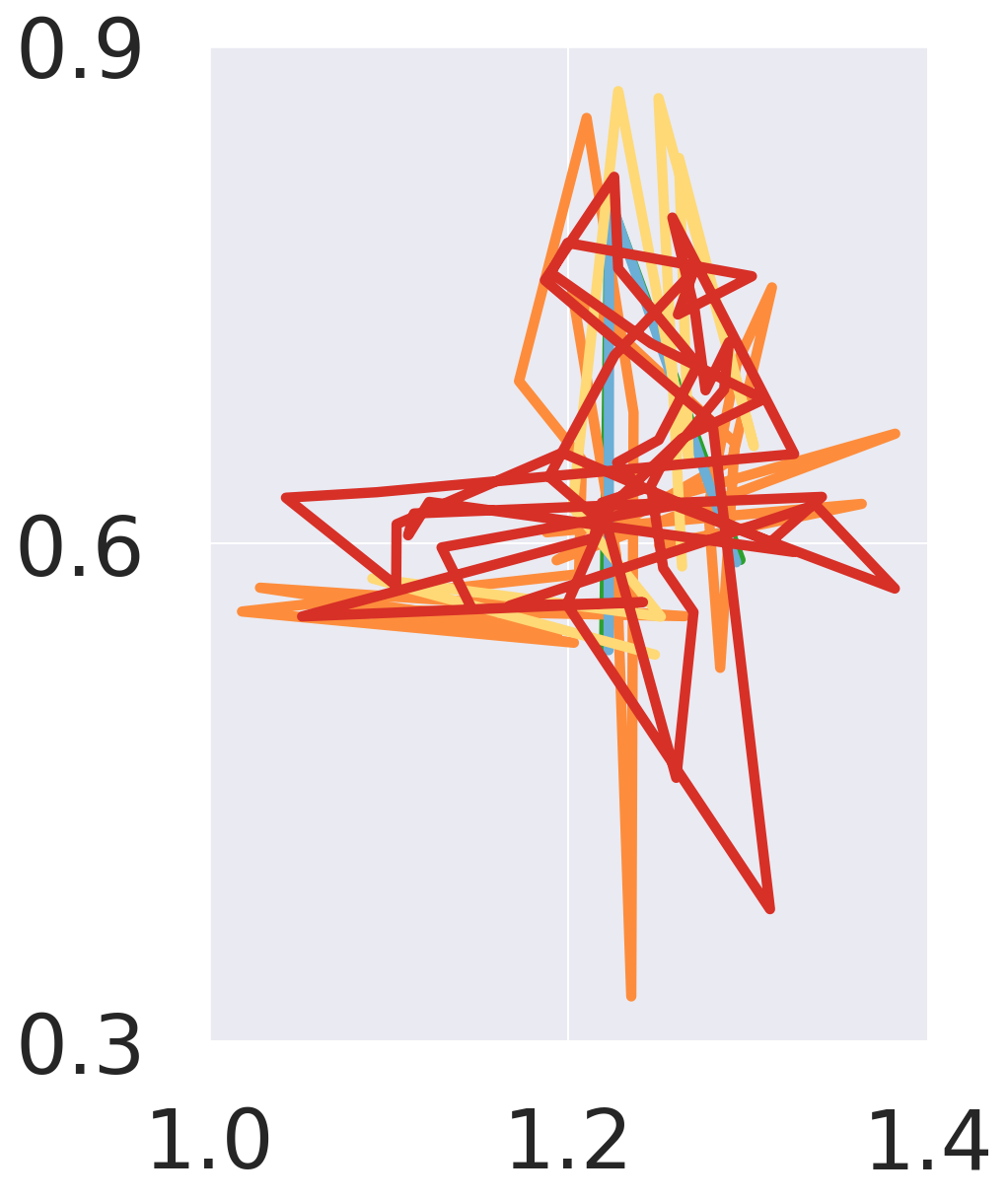} &
\includegraphics[width=0.30\columnwidth]{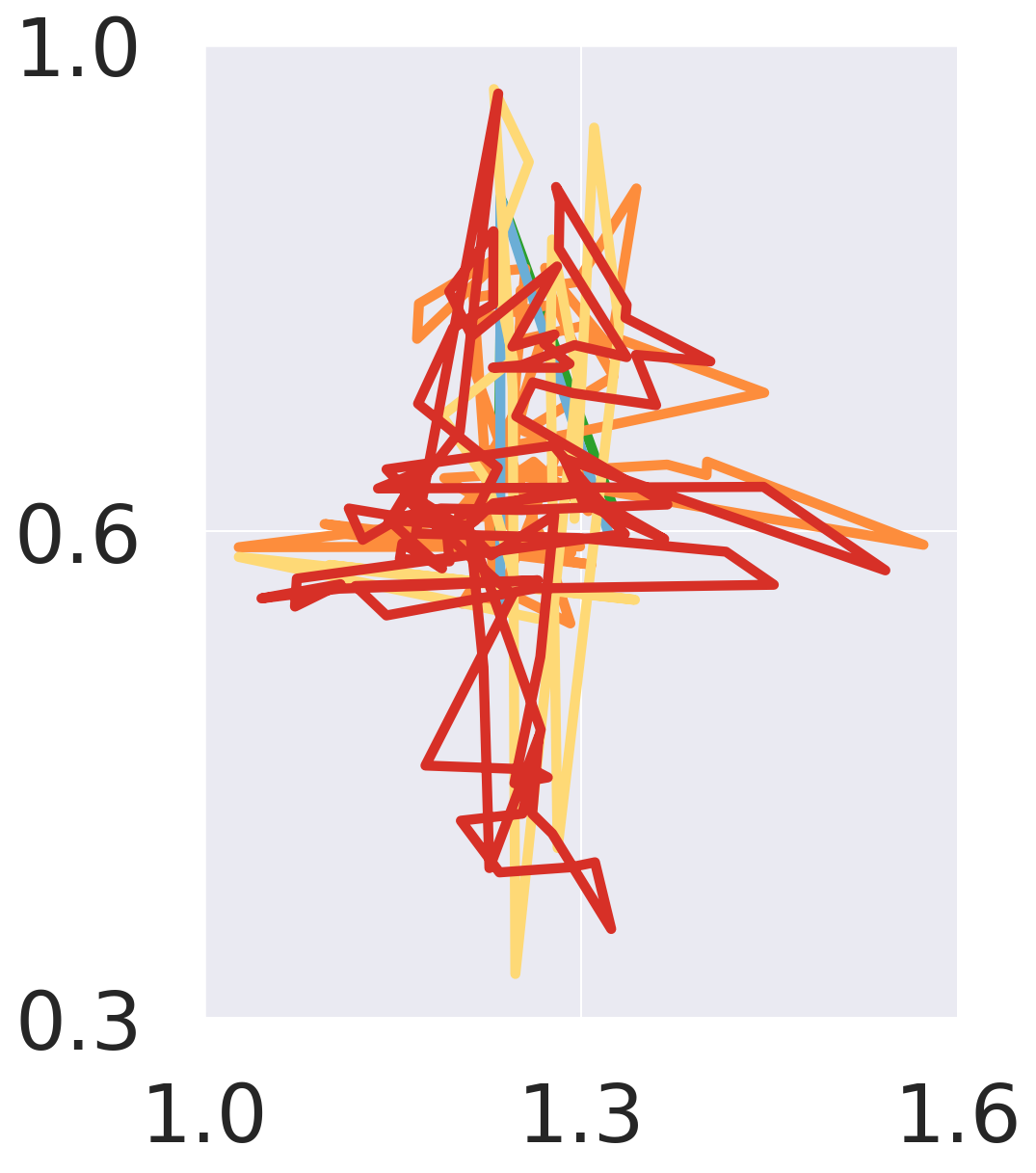}
\\[0pt]
\rotatebox{90}{\parbox{1.2cm}{\centering\scriptsize\textbf{DROID}}} &
\includegraphics[width=0.30\columnwidth]{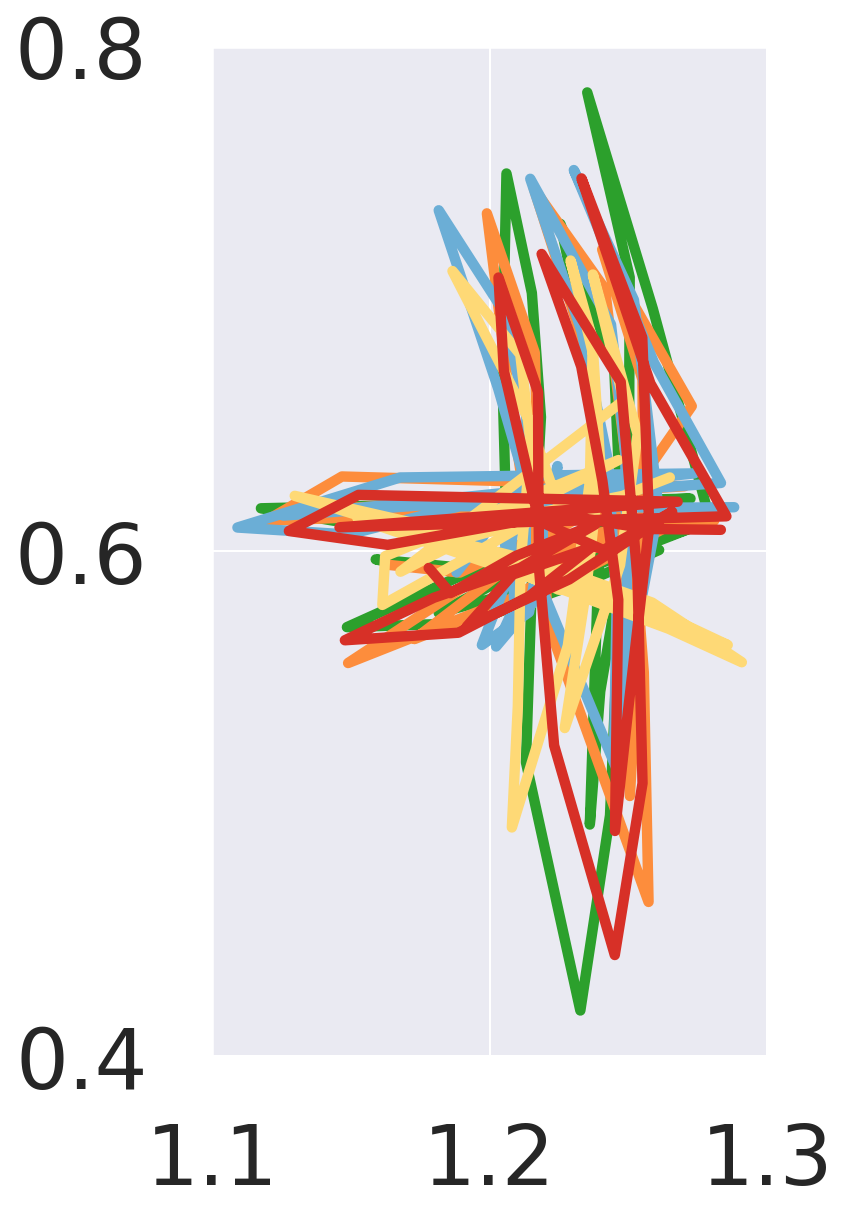} &
\includegraphics[width=0.30\columnwidth]{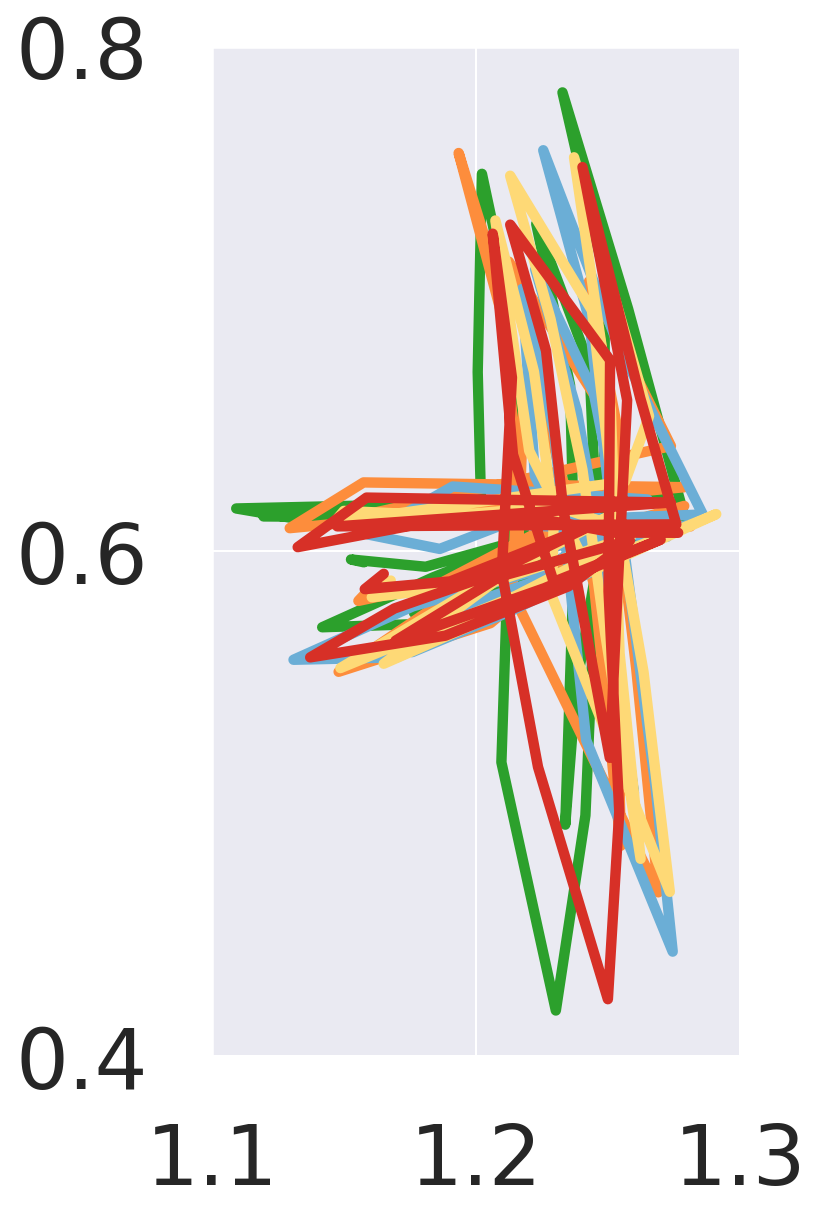} &
\includegraphics[width=0.30\columnwidth]{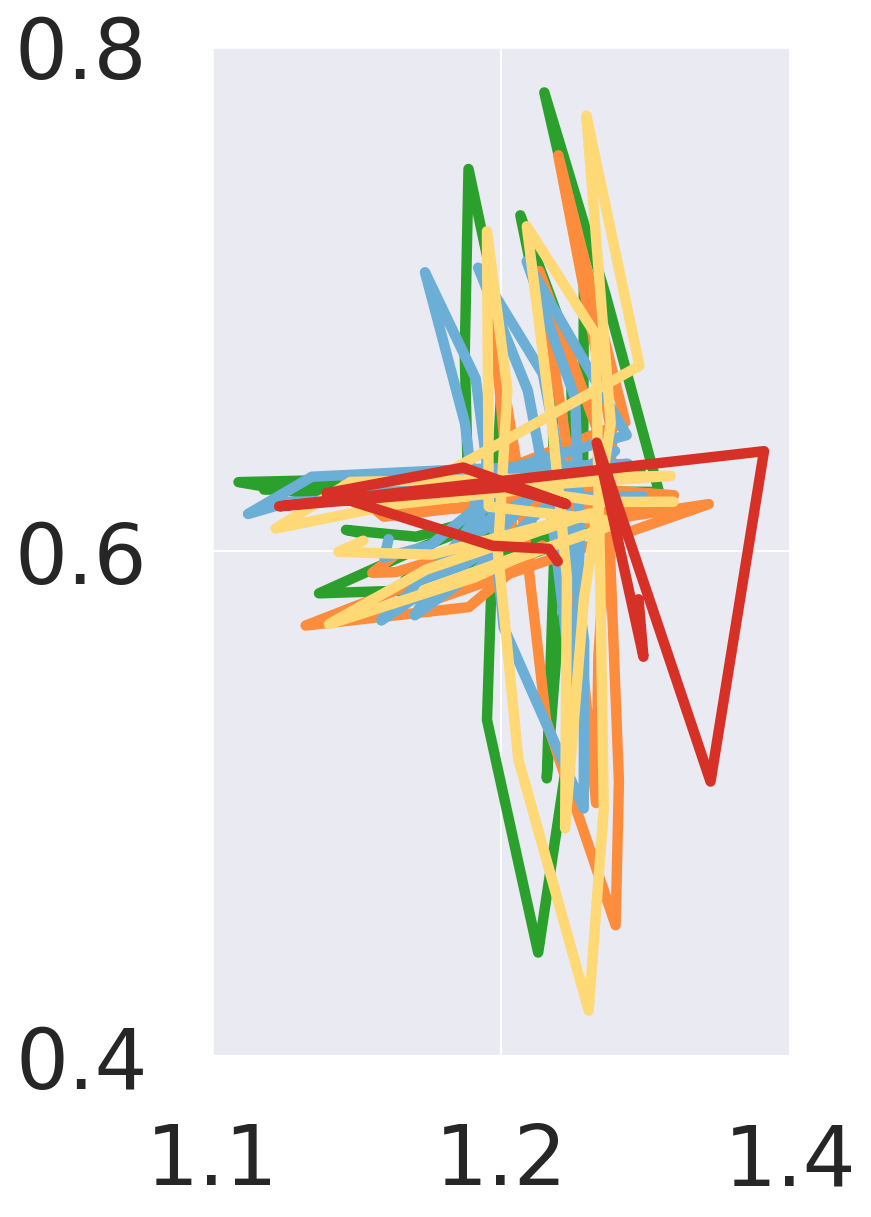} \\
\end{tabular}
\end{minipage}%
\begin{minipage}[t]{0.45\columnwidth}
\centering
\begin{tabular}{@{}c cc@{}}
 & \scriptsize\textbf{Bandwidth} & \scriptsize\textbf{Frame Drop} \\[0pt]
\rotatebox{90}{\parbox{1.2cm}{\centering\scriptsize\textbf{ORB3}}} &
\includegraphics[width=0.42\columnwidth]{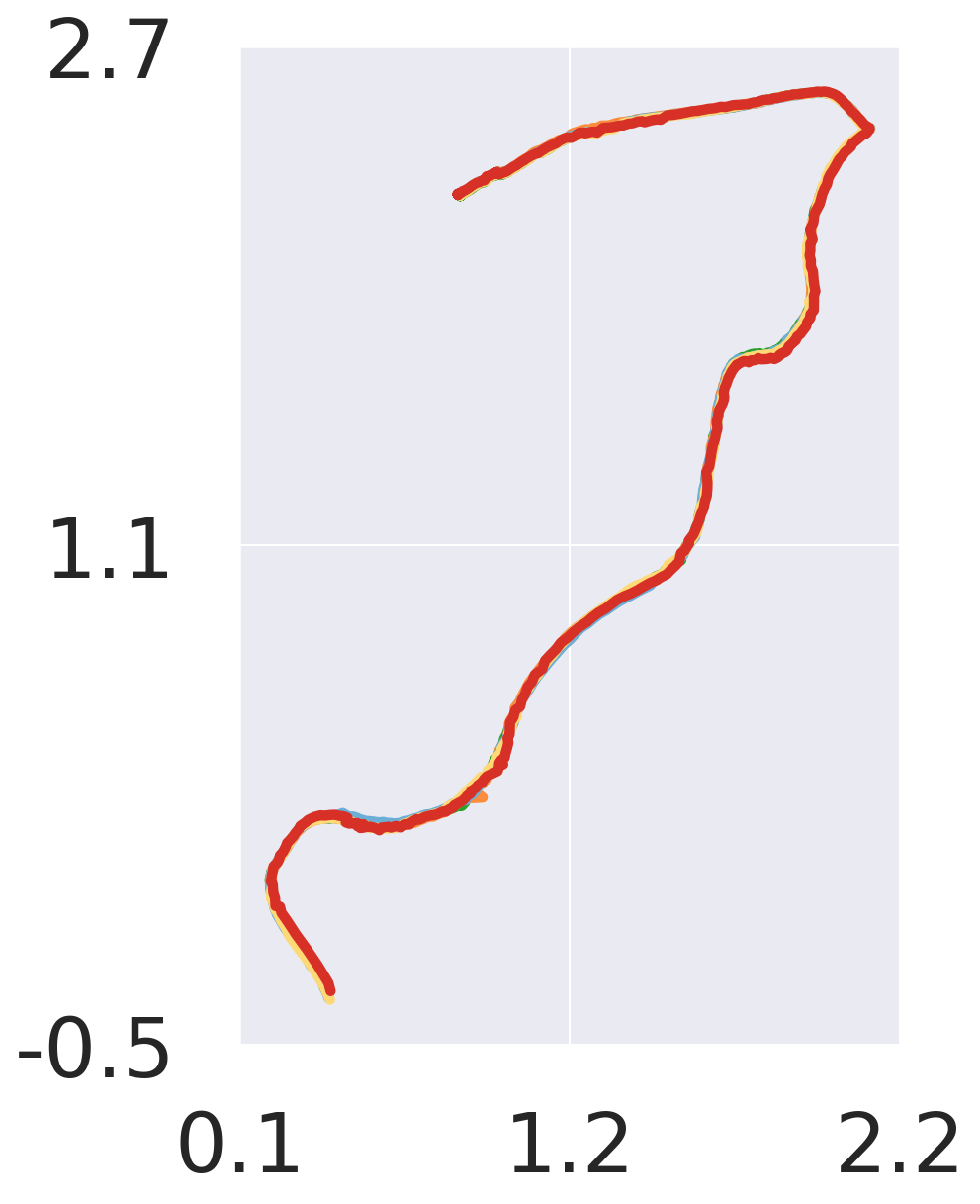} &
\includegraphics[width=0.42\columnwidth]{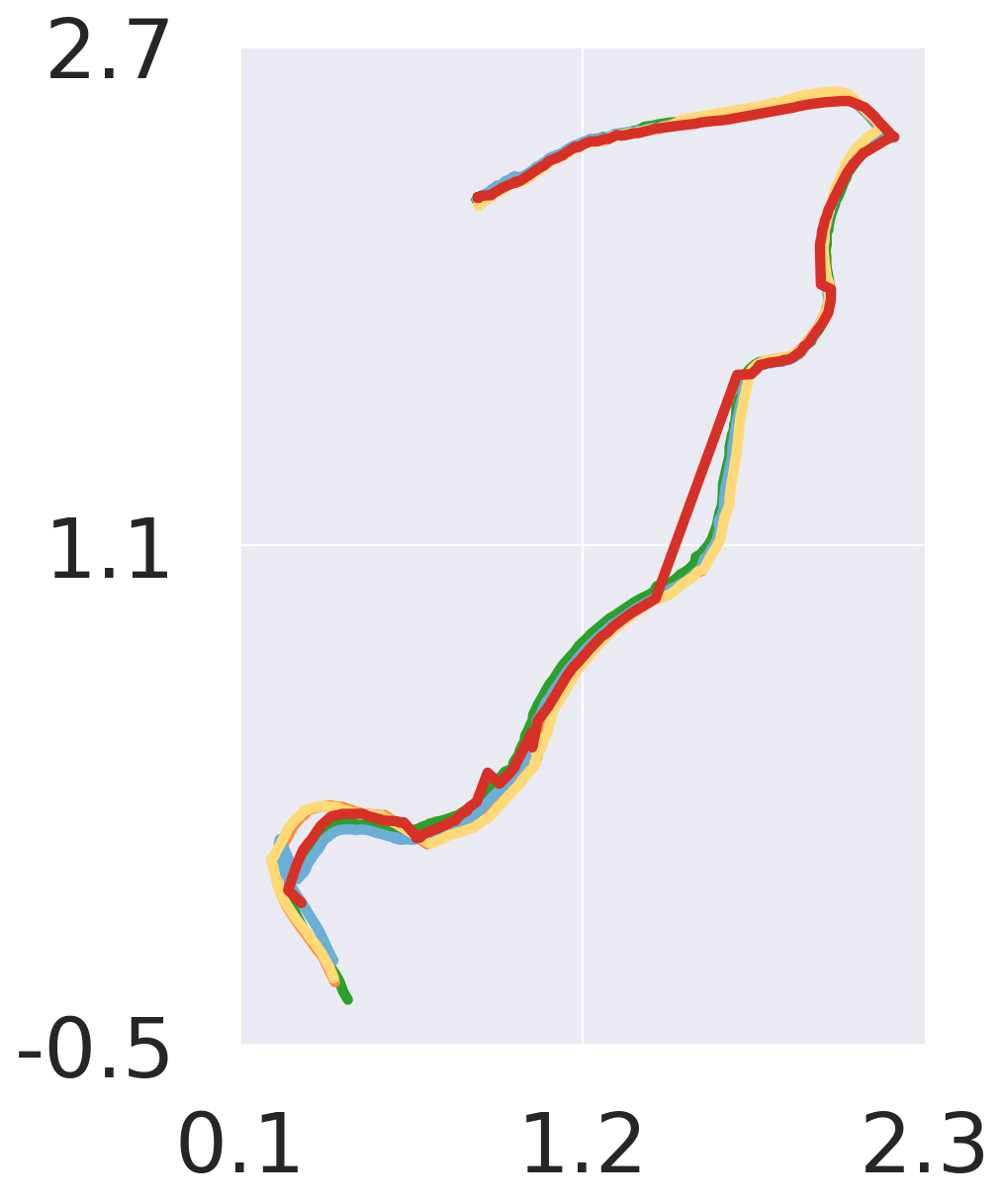}
\\[0pt]
\rotatebox{90}{\parbox{1.2cm}{\centering\scriptsize\textbf{VGGT}}} &
\includegraphics[width=0.42\columnwidth]{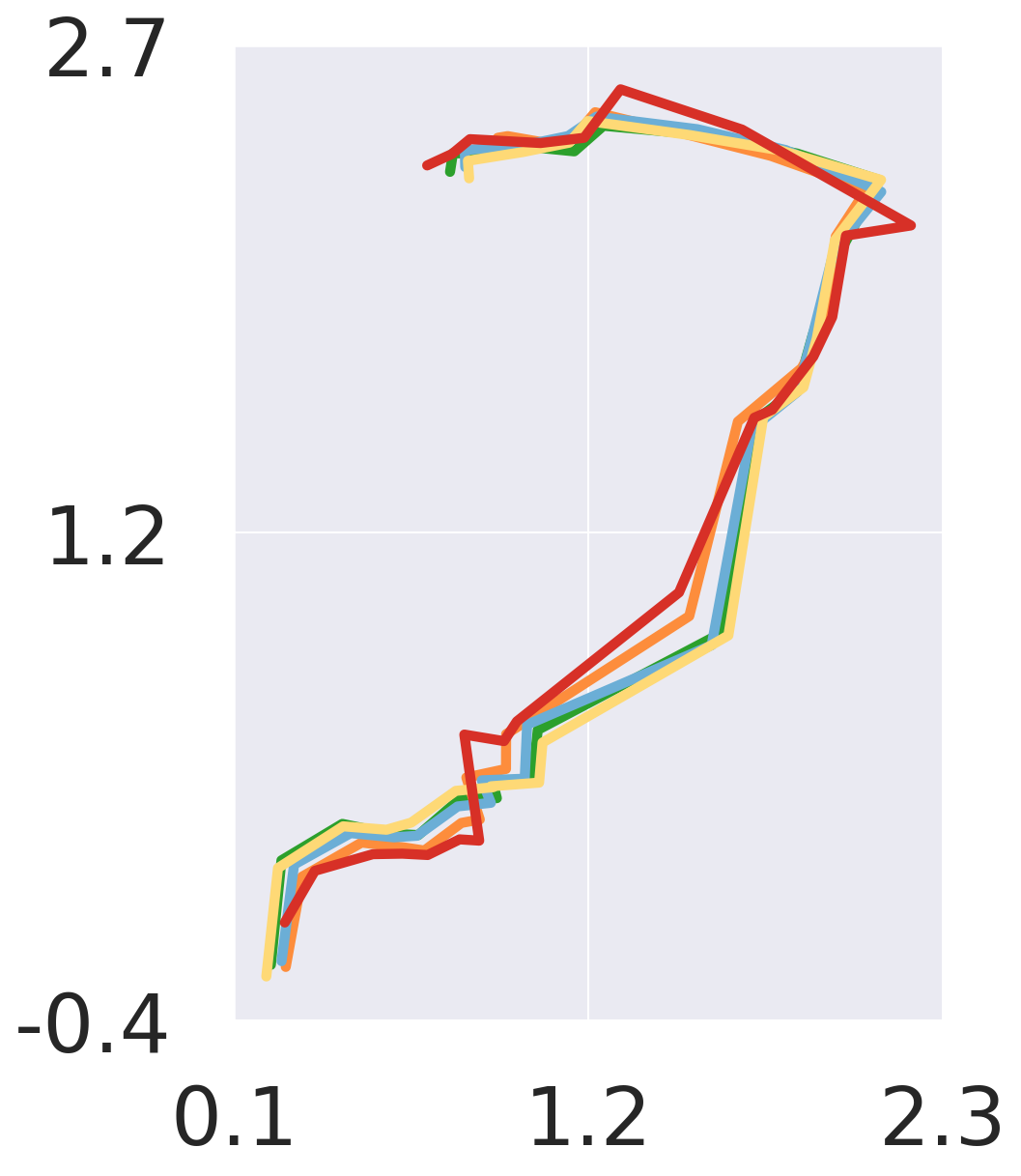} &
\includegraphics[width=0.42\columnwidth]{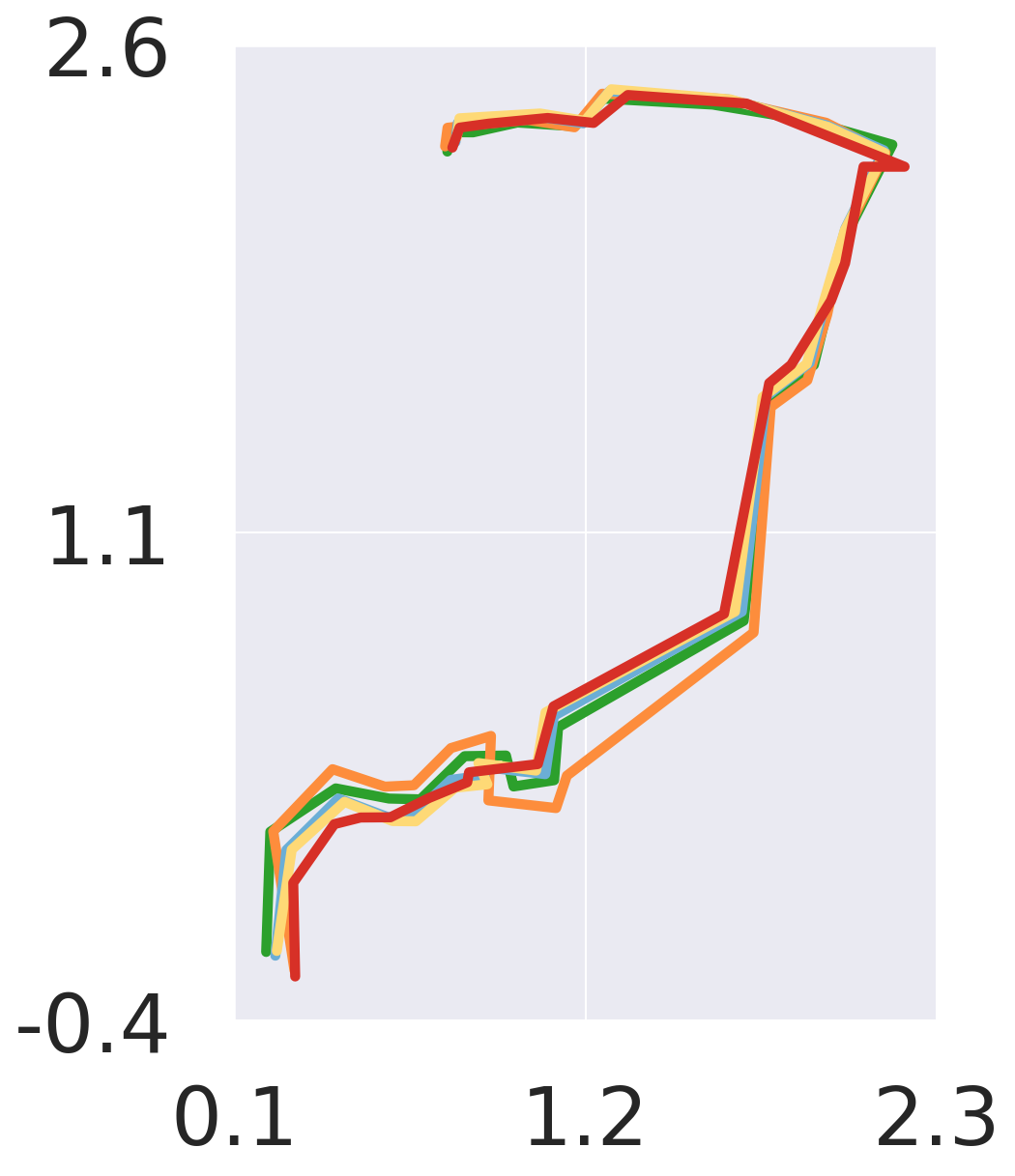}
\\[0pt]
\rotatebox{90}{\parbox{1.2cm}{\centering\scriptsize\textbf{Photo}}} &
\includegraphics[width=0.42\columnwidth]{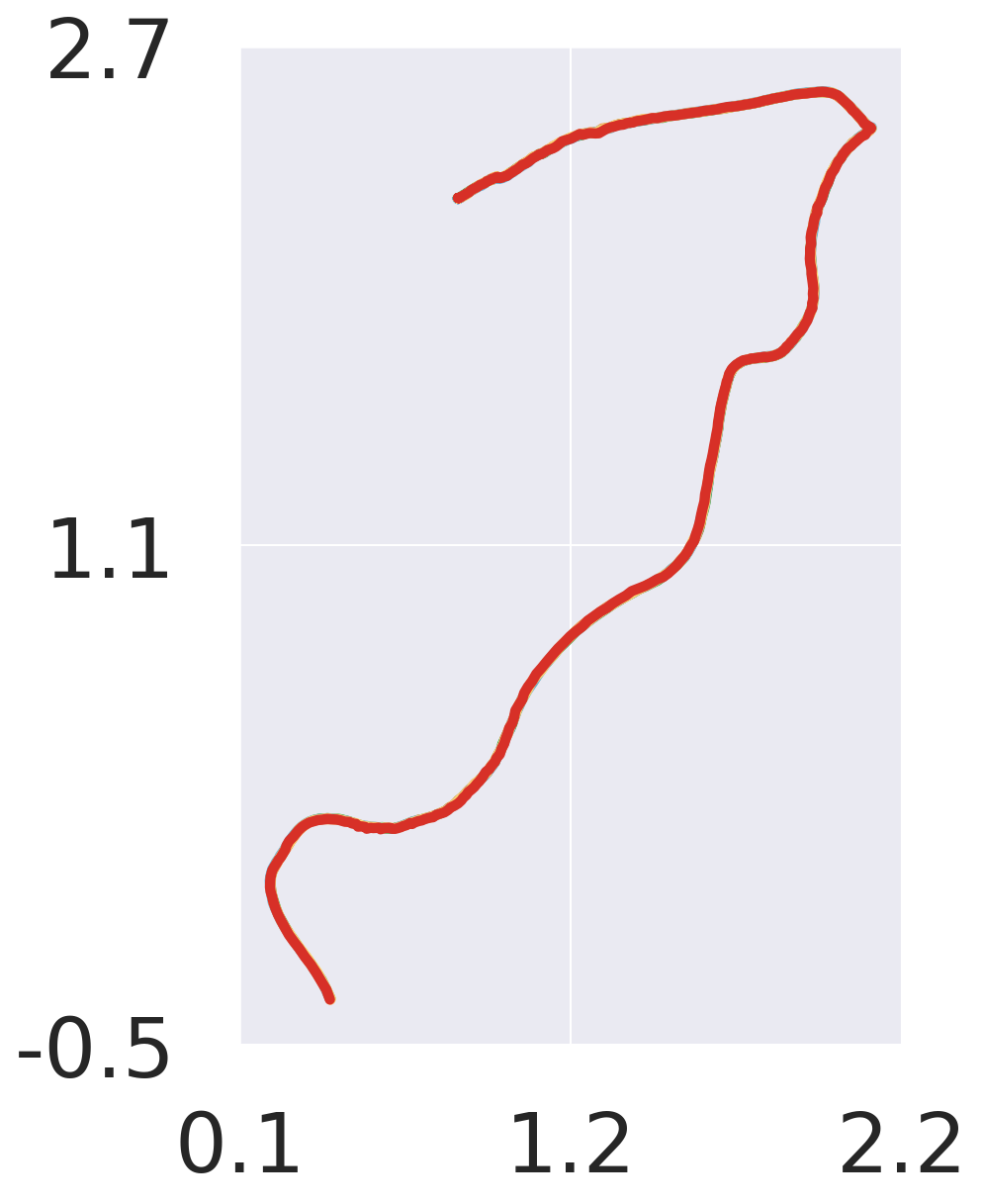} &
\includegraphics[width=0.42\columnwidth]{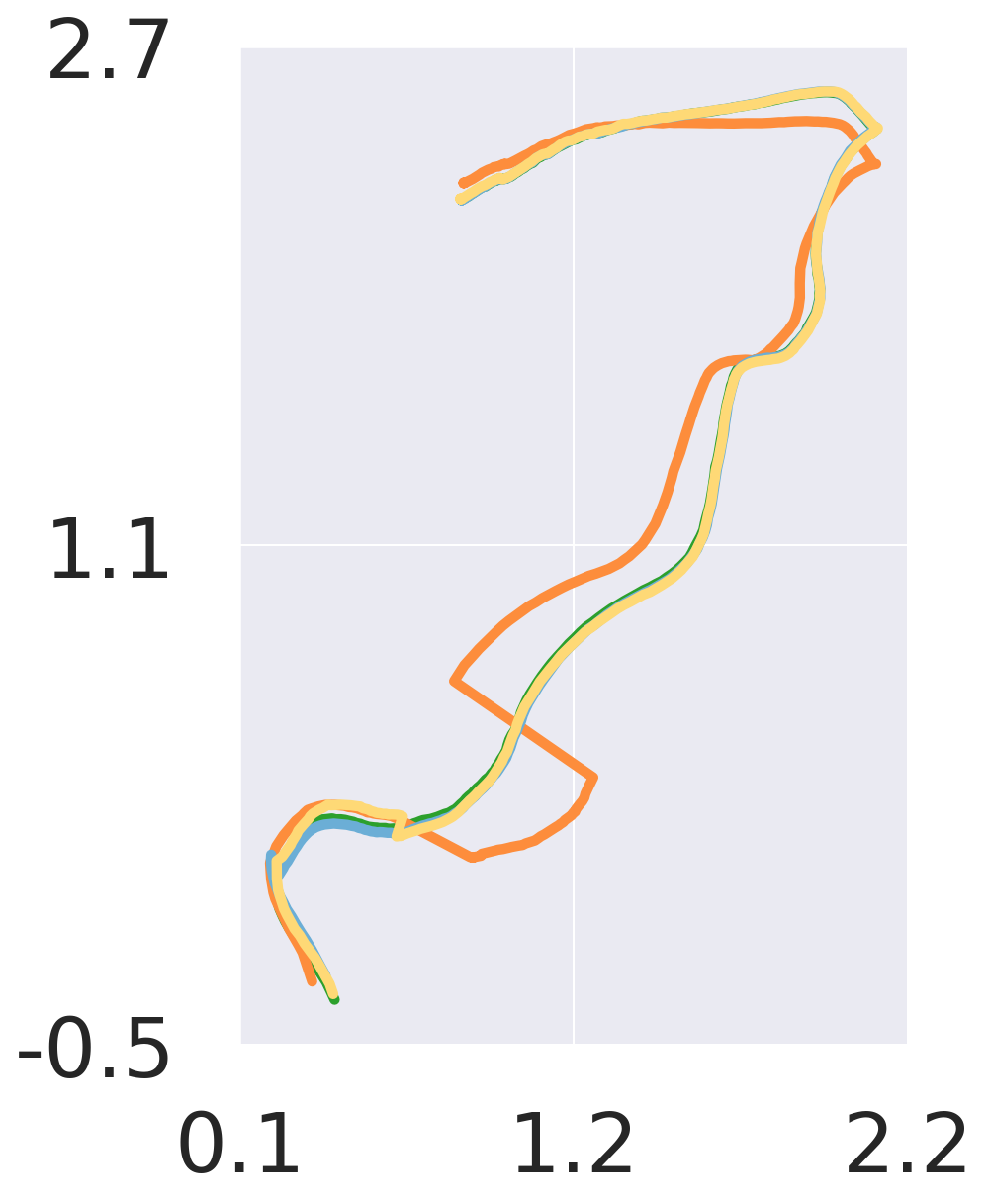} \\
\end{tabular}
\end{minipage}
\caption{Indoor (TUM left, EuRoC right) SLAM trajectory evaluation in the X-Y plane. Legend as in Fig.~\ref{fig:slam_results_kitti}.}
\label{fig:slam_results_indoor}
\end{figure}

\subsection{Results}
\label{subsec:slam_results}

\noindent We report trajectory accuracy, first for outdoor KITTI and then for indoor
TUM and EuRoC (Table~\ref{tab:results} and
Figures~\ref{fig:slam_results_kitti}--\ref{fig:slam_results_indoor}),
followed by feature tracking analysis (Table~\ref{tab:vo_results}) and
robustness boundary results (Table~\ref{tab:boundary_results}).


\subsubsection{Outdoor Trajectory Evaluation (KITTI)}

\noindent Table~\ref{tab:results} and Fig.~\ref{fig:slam_results_kitti} show outdoor trajectory results. Single-perturbation results show distinct sensitivities. Under rain, GigaSLAM
shows the largest degradation, while S3PO-GS changes least.
Under soiling, ORB-SLAM3 and GigaSLAM remain near baseline, whereas S3PO-GS
increases more. Composite perturbations show stronger effects. Night+Fog yields large errors
for all algorithms: ORB-SLAM3 reaches 30.26 at heavy and fails at severe,
while S3PO-GS and GigaSLAM both exceed $\Delta$+600. Rain+Blur also degrades
all algorithms, with GigaSLAM most affected ($\Delta$+2878). For
Rain+Soiling, the combined degradation exceeds either individual effect
for all algorithms. Overall, the feature-based ORB-SLAM3 is relatively more stable
than other algorithms under outdoor experiments but degrades sharply under Night+Fog.

\subsubsection{Indoor Trajectory Evaluation (TUM + EuRoC)}

\noindent Table~\ref{tab:results} and Fig.~\ref{fig:slam_results_indoor} show indoor trajectory results.
TUM results show ORB-SLAM3 and MASt3R-SLAM are more sensitive to cracking
than to soiling. For both, Soiling+Crack produces errors comparable to
cracking alone, suggesting that crack is the dominant factor in the combined
perturbation. DROID-SLAM
shows minimal degradation under all three conditions, with errors remaining
close to its baseline throughout. However, its baseline error is roughly
$10\times$ higher than ORB-SLAM3's. For EuRoC, we evaluate video-transport
perturbations. Bandwidth compression has less effect on ORB-SLAM3 and
Photo-SLAM than VGGT-SLAM.
Frame drop shows varied responses: Photo-SLAM degrades most, failing at
severe, while ORB-SLAM3 degrades moderately and VGGT-SLAM changes less with
frame drop but starts from a higher baseline ATE than ORB-SLAM3.

\begin{table}[!t]
\centering
\caption{SLAM results (ATE RMSE, m). $\Delta$: \% change from Clean to S (or H if S fails). \textcolor{red}{$\times$}: tracking failure. Clean is shared per algorithm/dataset.}
\label{tab:results}
\footnotesize
\setlength{\tabcolsep}{1.5pt}
\renewcommand{\arraystretch}{0.9}
\begin{tabular}{@{}l>{\raggedright\arraybackslash}p{1.6cm}|ccccc|r@{}}
\toprule
\textbf{Algo.} & \textbf{Perturbation} & \textbf{Clean} & \textbf{L} &
\textbf{M} & \textbf{H} & \textbf{S} & \textbf{$\Delta$\%} \\
\midrule
\multicolumn{8}{@{}l}{\scriptsize \textbf{KITTI-07} (200 frames, outdoor), ORB3: stereo, S3PO/Giga: mono} \\
\midrule
\multirow{5}{*}{ORB3} & Rain & 0.1326 & 0.1376 & 0.1514 & 0.1687 & 0.1738 & +31 \\
 & Soiling & 0.1326 & 0.1432 & 0.1419 & 0.1422 & 0.1339 & +1 \\
 & Rain+Soiling & 0.1326 & 0.1562 & 0.1526 & 0.1473 & 0.1923 & +45 \\
 & Night+Fog & 0.1326 & 0.1933 & 0.2081 & 30.26 & \textcolor{red}{$\times$} &
 +22k \\
 & Rain+Blur & 0.1326 & 0.1586 & 0.1763 & 0.2326 & 0.3623 & +173 \\
\midrule
\multirow{5}{*}{S3PO} & Rain & 1.0965 & 0.9923 & 1.1151 & 1.1805 & 1.3516 & +23 \\
 & Soiling & 1.0965 & 0.9978 & 0.9329 & 1.3266 & 1.4623 & +33 \\
 & Rain+Soiling & 1.0965 & 1.0204 & 1.0883 & 1.5027 & 3.7615 & +243 \\
 & Night+Fog & 1.0965 & 8.9032 & 9.1806 & 11.638 & 8.5446 & +679 \\
 & Rain+Blur & 1.0965 & 1.0442 & 1.2916 & 1.6603 & 3.5059 & +220 \\
\midrule
\multirow{5}{*}{Giga} & Rain & 0.2749 & 0.2361 & 0.2589 & 0.3268 & 0.7949 & +189 \\
 & Soiling & 0.2749 & 0.2572 & 0.2799 & 0.2674 & 0.2803 & +2 \\
 & Rain+Soiling & 0.2749 & 0.3085 & 0.2713 & 0.3285 & 1.0529 & +283 \\
 & Night+Fog & 0.2749 & 1.0334 & 1.4437 & 3.1060 & 3.1501 & +1046 \\
 & Rain+Blur & 0.2749 & 0.3486 & 0.3204 & 0.4258 & 8.1855 & +2878 \\
\midrule
\multicolumn{8}{@{}l}{\scriptsize \textbf{TUM fr1\_xyz} (all frames, indoor), ORB3: RGB-D, MASt3R/DROID: mono} \\
\midrule
\multirow{3}{*}{ORB3} & Soiling & 0.0103 & 0.0104 & 0.0105 & 0.0116 & 0.0602 &
+484 \\
 & Crack & 0.0103 & 0.0103 & 0.0811 & 0.1394 & 0.1714 & +1564 \\
 & Soiling+Crack & 0.0103 & 0.0104 & 0.1759 & 0.1701 & 0.1772 & +1620 \\
\midrule
\multirow{3}{*}{MASt3R} & Soiling & 0.0165 & 0.0179 & 0.0175 & 0.0312 & 0.0224 &
+36 \\
 & Crack & 0.0165 & 0.0194 & 0.0460 & 0.0799 & 0.0877 & +432 \\
 & Soiling+Crack & 0.0165 & 0.0240 & 0.0450 & 0.0850 & 0.1045 & +533 \\
\midrule
\multirow{3}{*}{DROID} & Soiling & 0.1143 & 0.1208 & 0.1285 & 0.0963 & 0.1211 &
+6 \\
 & Crack & 0.1143 & 0.1193 & 0.1021 & 0.0989 & 0.1087 & -5 \\
 & Soiling+Crack & 0.1143 & 0.1127 & 0.0914 & 0.1209 & 0.0969 & -15 \\
\midrule
\multicolumn{8}{@{}l}{\scriptsize \textbf{EuRoC V1\_01} (500 frames, indoor), ORB3/Photo: stereo, VGGT: mono} \\
\midrule
\multirow{2}{*}{ORB3} & Bandwidth & 0.0723 & 0.0752 & 0.0745 & 0.0777 &
0.0775 & +7 \\
 & Frame Drop & 0.0723 & 0.0772 & 0.1208 & 0.0740 & 0.0985 & +36 \\
\midrule
\multirow{2}{*}{VGGT} & Bandwidth & 0.1762 & 0.1619 & 0.1660 & 0.1630 &
0.2087 & +18 \\
 & Frame Drop & 0.1762 & 0.1830 & 0.1780 & 0.1738 & 0.1944 & +10 \\
\midrule
\multirow{2}{*}{Photo} & Bandwidth & 0.0725 & 0.0724 & 0.0724 & 0.0723 &
0.0722 & 0 \\
 & Frame Drop & 0.0725 & 0.0793 & 0.0824 & 0.1607 & \textcolor{red}{$\times$} & +122 \\
\bottomrule
\end{tabular}
\end{table}

\begin{table}[h]
\centering
\vspace*{2mm}
\caption{Feature tracking results (mean track length, frames). $\Delta$: \% change from C to S (or H if S fails). \textcolor{red}{$\times$}: feature-matching failure.}
\label{tab:vo_results}
\setlength{\tabcolsep}{1.5pt}
\footnotesize
\renewcommand{\arraystretch}{0.9}
\begin{tabular}{@{}ll>{\raggedright\arraybackslash}p{1.0cm}|ccccc|r@{}}
\toprule
\textbf{Dataset} & \textbf{Feature} & \textbf{Perturb.} & \textbf{Clean} & \textbf{L} &
\textbf{M} & \textbf{H} & \textbf{S} & \textbf{$\Delta$\%} \\
\midrule
\multirow{2}{*}{KITTI} & ORB & Rain & 8.24 & 8.30 & 8.44 & 8.07 & \textcolor{red}{$\times$} & $-$2 \\
 & SuperPoint & Rain & 7.85 & 7.71 & 7.56 & 7.24 & 5.19 & $-$34 \\
\midrule
\multirow{2}{*}{TUM} & ORB & Soiling & 106.23 & 112.74 & 120.42 & 121.93 & 120.73 & +14 \\
 & SuperPoint & Soiling & 53.35 & 65.82 & 110.82 & 86.53 & 58.78 & +10 \\
\midrule
\multirow{2}{*}{EuRoC} & ORB & Band. & 20.74 & 17.58 & 15.73 & 18.91 & \textcolor{red}{$\times$} & $-$9 \\
 & SuperPoint & Band. & 16.95 & 16.79 & 15.89 & 15.80 & 16.34 & $-$4 \\
\bottomrule
\end{tabular}
\end{table}

\begin{table}[h]
\centering
\caption{Robustness-boundary results. Each fail/pass side reports the parameter value (in the units of Range) and its ATE RMSE. $N$ = trials.}
\label{tab:boundary_results}
\footnotesize
\setlength{\tabcolsep}{2.5pt}
\renewcommand{\arraystretch}{0.9}
\begin{tabular*}{\columnwidth}{@{\extracolsep{\fill}}l c l c c>{\raggedright\arraybackslash}p{2.45cm}@{}}
\toprule
\textbf{Dataset} & \textbf{Algo.} & \textbf{Perturb.} & \textbf{Range} & \textbf{N} & \textbf{Fail/Pass (val, ATE)} \\
\midrule
KITTI & ORB3 & Night+Fog & 10--200\,m & 8 & 21,28.255 / 24,0.304 \\
EuRoC & Photo & Frame Drop & 10--50\% & 6 & 47,\textcolor{red}{$\times$} / 45,0.219 \\
\bottomrule
\end{tabular*}
\end{table}

\subsubsection{Feature Tracking Analysis}

\noindent Table~\ref{tab:vo_results} reveals several patterns in feature tracking under
perturbations. First, track length is often non-monotonic: it does not
consistently decrease with intensity. For example, the mean ORB track length on
TUM increases under soiling even as we observe mean matched features dropping
from 207 to 115, likely because the result depends on where soiling falls in the
image, with regions containing weaker features being occluded. Second, failure
can be abrupt rather than
gradual. ORB maintains stable tracking through heavy rain but fails completely at severe,
with no intermediate degradation. Third, neither classical nor learned features
are universally better: ORB maintains longer tracks under clean conditions, while SuperPoint survives severe rain and severe bandwidth where ORB fails.

\subsubsection{Robustness Boundary Results}
\label{subsec:boundary_results}

\noindent Table~\ref{tab:boundary_results} summarizes robustness-boundary search outcomes
for two configurations that exhibited tracking failures in the severity sweeps.
For ORB-SLAM3 on KITTI under night+fog, the boundary lies between 21\,m and
24\,m visibility: ATE jumps from 0.304\,m to 28.255\,m over just 3\,m of
visibility change. For Photo-SLAM on EuRoC under frame drop, the boundary lies
between 45\% and 47\% drop rate, where ATE rises from 0.219\,m to a tracking
failure. Across these two searches, KITTI night+fog required 8 trials,
compared with 37 if evaluated by an incremental sweep, and EuRoC frame drop required 6 trials vs 14,
corresponding to ~4.6x and ~2.3x runtime reductions. These results show that boundary search identifies failure boundaries
efficiently.


\section{Conclusion}
\label{sec:conclusion}

\noindent We presented SAL, a modular framework for evaluating visual SLAM robustness
under adversarial conditions. SAL defines perturbation modules parameterized in real-world units with
depth-aware and composite effects. It provides an extensible architecture for datasets,
perturbations, and SLAM systems. To complement
severity sweeps, SAL introduces robustness-boundary search to find failure thresholds. Together, these capabilities make robustness
evaluation more realistic, extensible, and efficient. We demonstrated capabilities across KITTI,
TUM, and EuRoC with multiple perturbation families and seven SLAM algorithms.
SAL is open-sourced at \url{https://github.com/sfu-rsl/SLAMAdversarialLab}.

\bibliographystyle{IEEEtran}
\bibliography{references}

\end{document}